%% file: main.tex
\documentclass[twoside]{article}
\pdfoutput=1
\usepackage{aistats2020}
\usepackage[utf8]{inputenc} 
\usepackage[T1]{fontenc}    
\usepackage{hyperref}       
\usepackage{url}            
\usepackage{booktabs}       
\usepackage{amsfonts}       
\usepackage{nicefrac}       
\usepackage{microtype}      

\usepackage{caption}
\usepackage{calc}
\usepackage{amsmath, amssymb, amsthm}
\usepackage{parskip}
\usepackage{color,hyperref}
\usepackage{epsfig}
\usepackage{subfig}
\usepackage{verbatim}
\usepackage{scalerel}
\usepackage{rotating}
\usepackage{enumitem}
\usepackage{wrapfig}
\usepackage{cleveref}
\usepackage{bbm}
\usepackage{array}
\usepackage{soul}
\usepackage{xspace}
\usepackage[numbers,sort,compress]{natbib}

\usepackage{graphicx}
\usepackage{comment}
\usepackage{algorithm}
\usepackage[noend]{algpseudocode}
\usepackage{setspace}

\usepackage{kky}
\usepackage{ot_dist_defns}

\begin{document}

%

%

\twocolumn[

\aistatstitle{ChemBO: Bayesian Optimization of Small Organic Molecules with Synthesizable
Recommendations}

\vspace{-2mm}
\aistatsauthor{
Ksenia Korovina$^1$ \hspace{10mm}
Sailun Xu$^1$ \hspace{10mm}
Kirthevasan Kandasamy$^2$ \hspace{10mm}
Willie Neiswanger$^1$
}
\vspace{2mm}

\aistatsauthor{
Barnab\'as P\'oczos$^1$ \hspace{10mm}
Jeff Schneider$^1$ \hspace{10mm}
Eric P.~Xing$^1$
}
\vspace{5mm}

\aistatsaddress{
\hspace{-15mm}
$^1$Carnegie Mellon University \hspace{5mm}
$^2$U. C. Berkeley}





]

\setlength{\parskip}{6.0pt plus 2.0pt}

\input{tables.tex}
\input{figures.tex}

\input{abstract.tex}

\input{introduction.tex}

\input{method.tex}

\input{experiments.tex}

\input{conclusion.tex}

\subsubsection*{Acknowledgments}
We would like to thank Christopher R.~Collins for reviewing the initial draft of this manuscript.


{\small
\renewcommand{\bibsection}{\section*{References} }
\setlength{\bibsep}{1.1pt}
\bibliography{main}
}
\bibliographystyle{unsrtnat}

\newpage

{\Large{\textbf{Appendix}}}

\appendix
\input{app_dist.tex}

\input{app_model_details.tex}
\input{app_experiments.tex}

\end{document}

%% file: tables.tex
\renewcommand{\algorithmicrequire}{\textbf{Input:}}
\renewcommand{\algorithmicensure}{\textbf{Output:}}
\algnewcommand{\algorithmicgoto}{\textbf{go to}}%
\algnewcommand{\Goto}[1]{\algorithmicgoto~\ref{#1}}%

\newcommand{\synthesize}{\textproc{Synthesize}}
\newcommand{\randselect}{\textproc{Rand-Select}}
\newcommand{\nullmol}{\textproc{Null}}

\newcommand{\insertAlgoRandExplorer}{

\begin{algorithm}[t]
\small
  \caption{\textproc{Acquisition-Opt}: Random Walk Explorer}\label{algo:explorer}
  \begin{algorithmic}[1]
  \State \textbf{Input:}   {$n$, $\mathcal{S}$, $\Pcal$, $D$}
\Comment{Steps $n$, Initial 
molecules $\mathcal{S}$ and conditions $\Pcal$, Past evaluations $D$}
  \State $k=0$
    \While{$k\leq n$} \label{marker}
      \State $S \gets \randselect(\mathcal{S})$\Comment{Select a subset of molecules as reaction inputs}
      \State $Q \gets \randselect(\Qcal)$ \Comment{Select a subset of process conditions}
      \State $M \gets \synthesize(S, Q)$ \Comment{Predict product}
    \If{$M\neq \nullmol\;$ \textbf{and} $\;M\backslash D \neq \emptyset$  }
      \Comment{$M\backslash D$ is set difference.}
      \State $k \gets k + 1$
      \State $\Scal \gets \Scal \cup M\backslash D$
           \Comment{Add outcomes to the pool}
    \EndIf
    \EndWhile
    \Return $\argmax_{x\in \mathcal{S}} \varphi(x)$
    \label{explorerreturn}
  \end{algorithmic}
\end{algorithm}

}

\newcommand{\insertVirtualScreeningTable}{
\begin{table}[H]
\centering
\resizebox{0.6\columnwidth}{!}{%
\begin{tabular}{|l|l|l|l|}
\hline
QED & penalized logp \\ \hline
$0.922 \pm 0.013$ & $5.34 \pm 0.973$ \\ \hline  
\end{tabular}
}
\caption{\small Virtual screening baseline: means and standard deviations over 10 replications.
\label{table:virt}
}
\vspace{-4mm}
\end{table}
}

\newcommand{\insertComparisonTable}{
\begin{table*}[t]
\centering
\resizebox{1.5\columnwidth}{!}{%
\begin{tabular}{|c|c|c|c|c|c|}
\hline
 & ORGAN~\citep{organ} & JT-VAE~\citep{jt-vae} & GCPN~\citep{gcpn} &
MolDQN~\citep{zhou2018optimization} & \Chem{} (ours) \\ \hline
QED & $ 0.896$ & $ 0.925$ & $ 0.948$ & $0.948$ & $0.941$ \\ \hline
Pen-logP & $3.63$ & $5.30$ & $7.98$ & $11.84$ & $18.39$ \\ \hline
\# evaluations & $\geq 5K$ & $275K$ & $\geq 25K$ & $\geq 25K$ & $100$ \\ \hline
\end{tabular}
}
\caption{\small The best QED and Pen-LogP scores reported from prior work.
For \Chem, we use the best value obtained across the 5 trials for both
\fingerprint{} and \dist.
Note that not all methods treat this as an optimization problem and
do not impose conditions on synthesizability as we do.
}
\vspace{-4mm}
\label{tab:compar_res}
\end{table*}
}

\newcommand{\insertSASTable}{
\begin{table}[H]
\centering
\resizebox{\columnwidth}{!}{%
\begin{tabular}{|l|l|l|l|}
\hline
ChEMBL & ZINC250k & Avg SA score & Min path SA \\ \hline
$2.73 \pm 0.65$ & $3.1 \pm 0.77$ & $3.77 \pm 1.46$ &  $2.5 \pm 0.44$ \\ \hline
\end{tabular}
}
\caption{\small Synthetic accessibility scores over 50 samples/runs over the datasets, optimal results from \Chem{}, and average minimum over produced synthesis paths. \label{table:sas_scores}}
\vspace{-4mm}
\end{table}
}

%% file: figures.tex
\newcommand{\insertFigOrganicEgs}{
\begin{figure}[H]
\begin{center}
\includegraphics[width=2.7in]{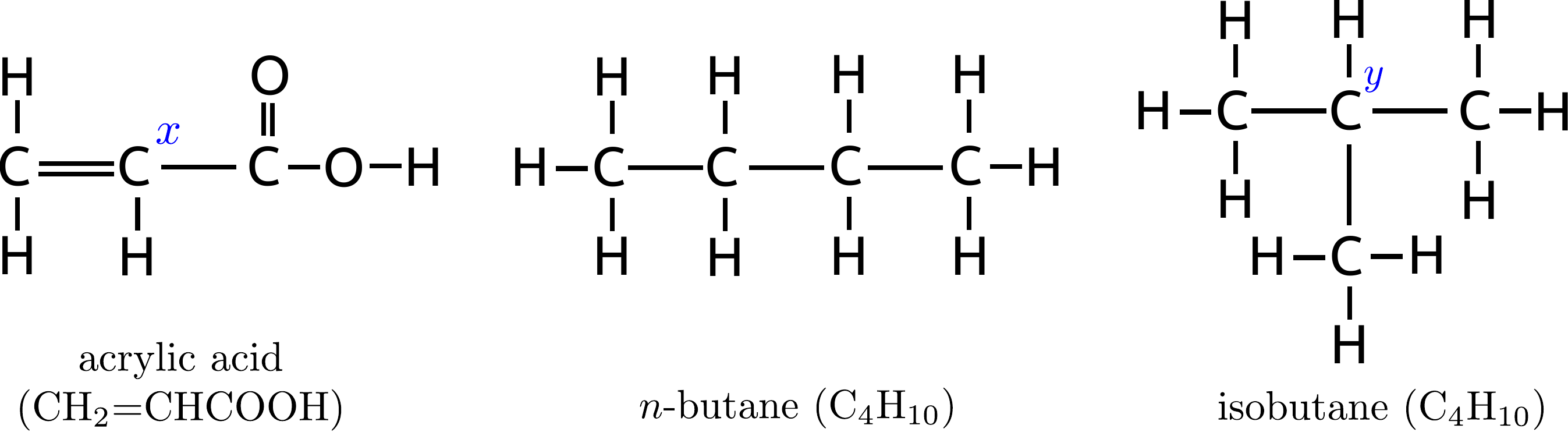}
\caption{{\small Examples of organic molecules}\label{fig:organicegs}}
\vspace{-2mm}
\end{center}
\end{figure}
}

\newcommand{\insertFigGlucose}{
\begin{center}
\includegraphics[width=3.2in]{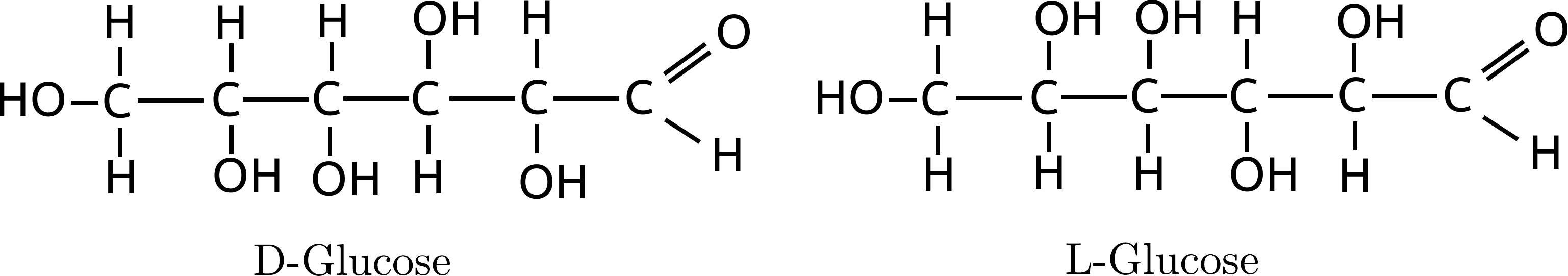}
\end{center}
}

\newcommand{\insertDataStats}{
\begin{figure*}[t]
\begin{tabular}{@{}c@{\!\!\!\!\!}c@{\!\!\!\!\!}c@{\!\!\!\!\!}c@{}}
\includegraphics[width=0.5\textwidth, clip=true, trim=6mm 0mm 6mm 0mm]{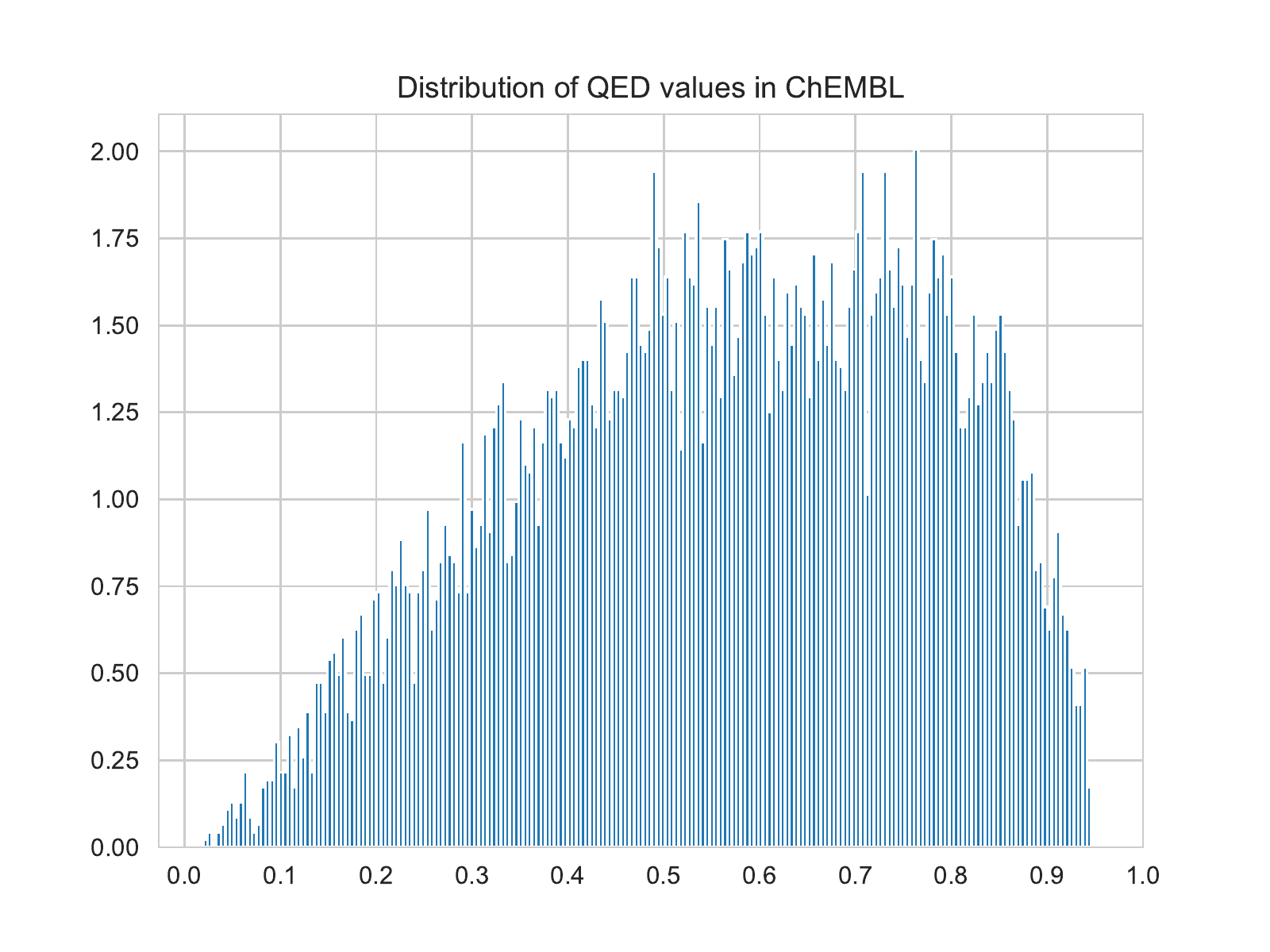}
&
\includegraphics[width=0.5\textwidth, clip=true, trim=6mm 0mm 6mm 0mm]{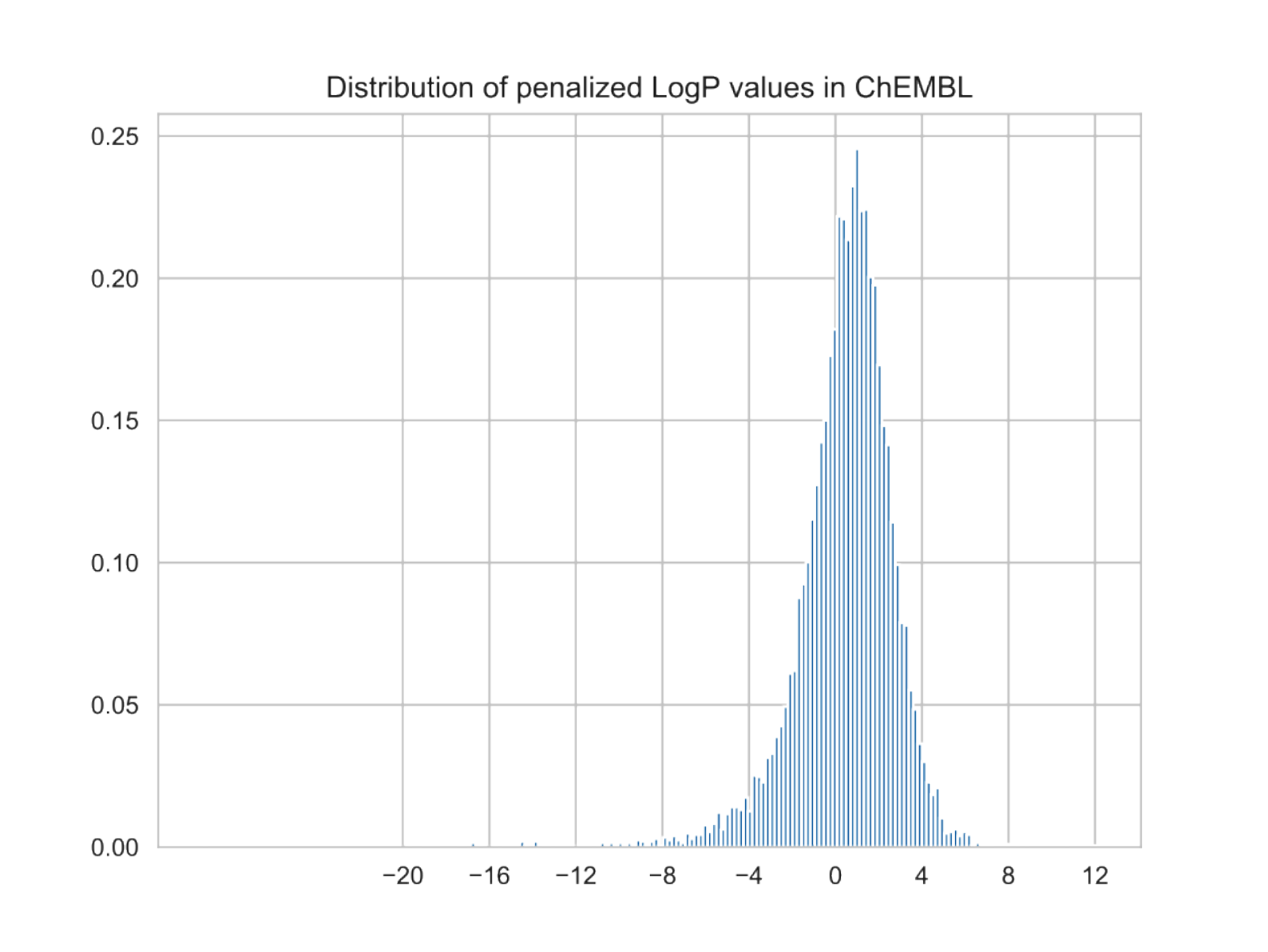}
\\[-1mm]
\end{tabular}
\caption{\small ChEMBL dataset statistics: normalized histograms of QED score and penalized logP score.}\label{fig:data_stats}
\end{figure*}
}

\newcommand{\insertTsneVis}{
\begin{figure*}[t]
\centering
\begin{tabular}{@{}c@{\!\!\!\!\!}c@{\!\!\!\!\!}c@{\!\!\!\!\!}c@{}}
\includegraphics[width=0.3\textwidth, clip=true, trim=6mm 0mm 6mm 0mm]{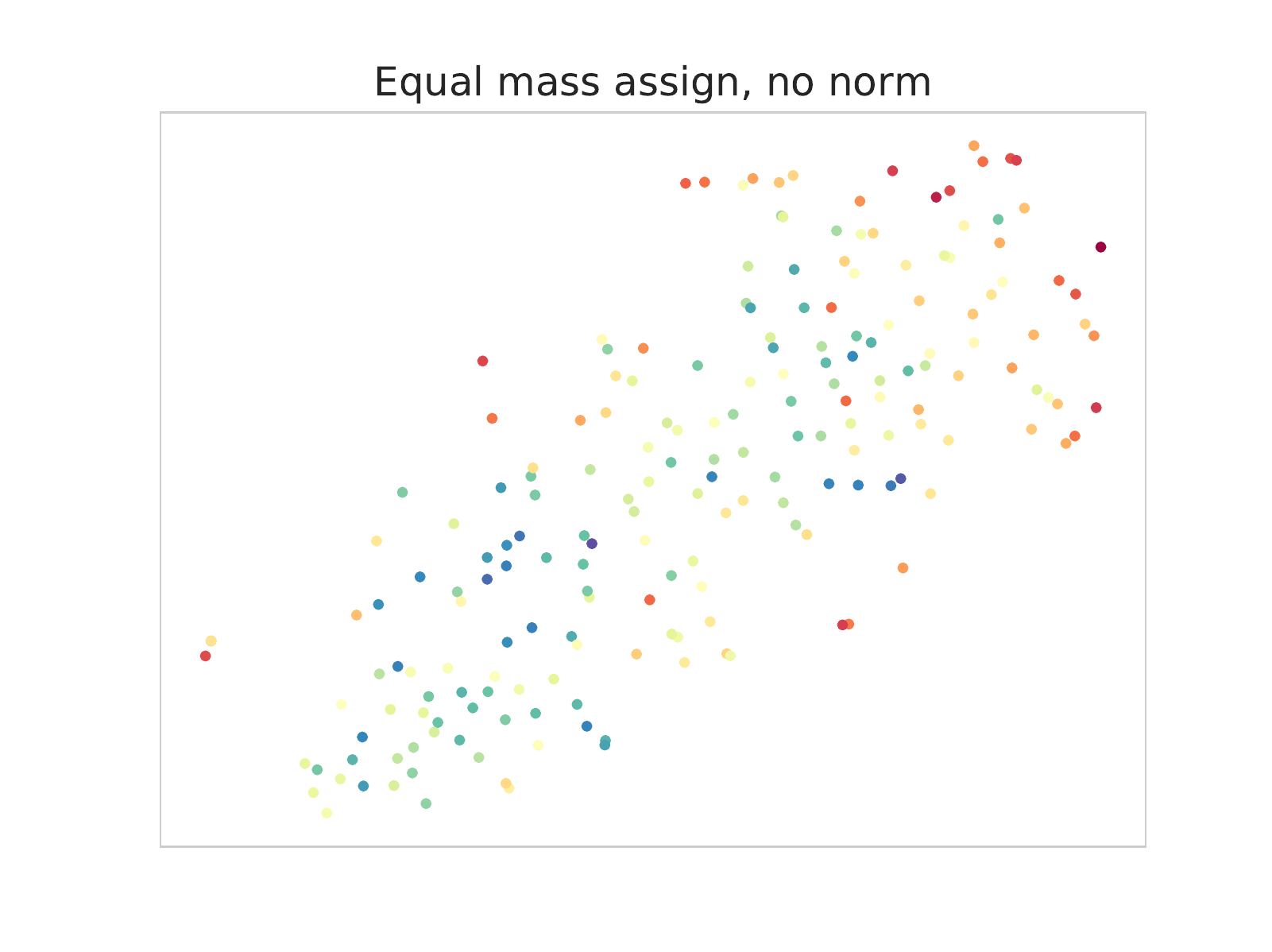}
&
\includegraphics[width=0.3\textwidth, clip=true, trim=6mm 0mm 6mm 0mm]{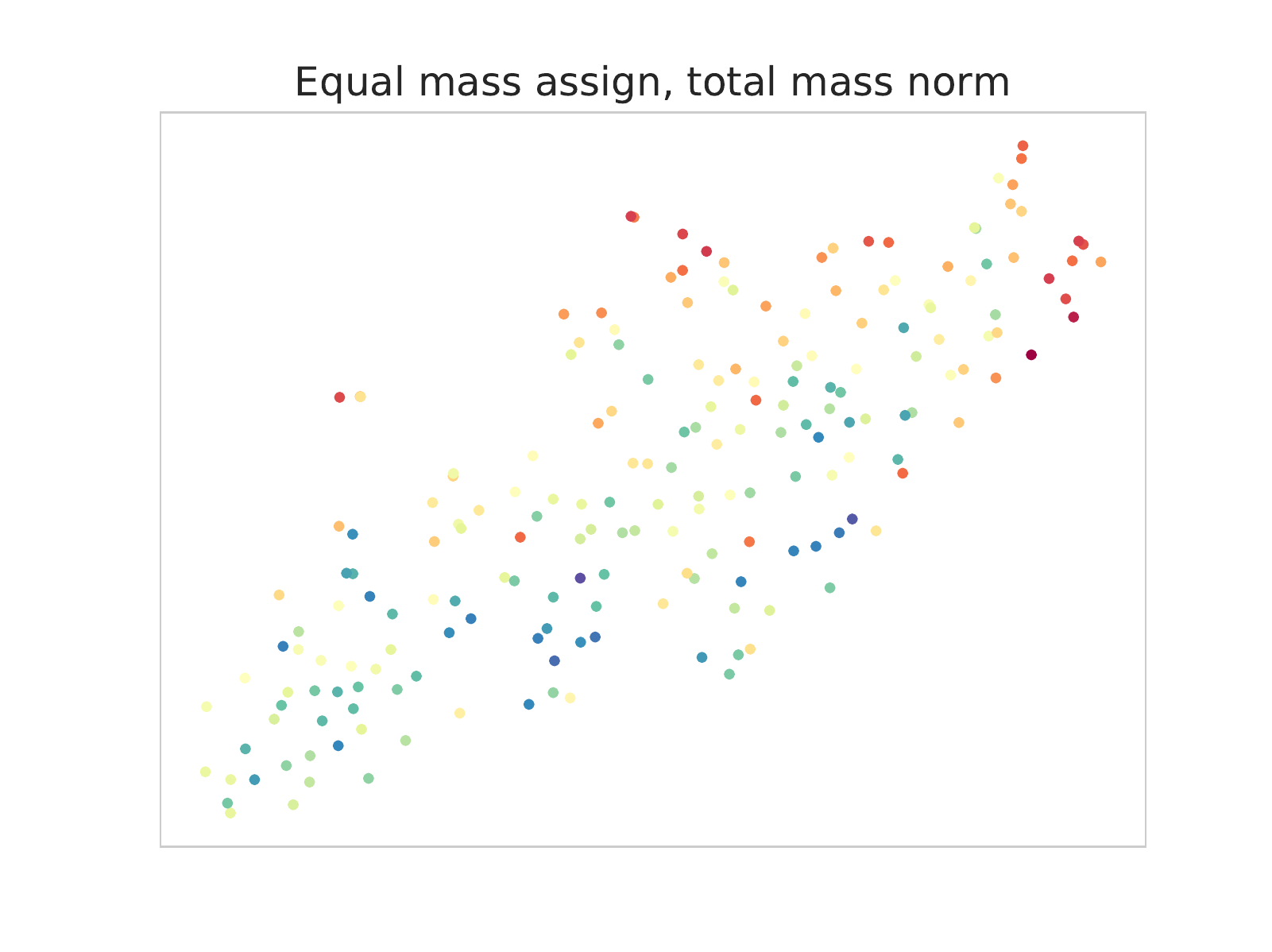}
\\
\includegraphics[width=0.3\textwidth, clip=true, trim=6mm 0mm 6mm 0mm]{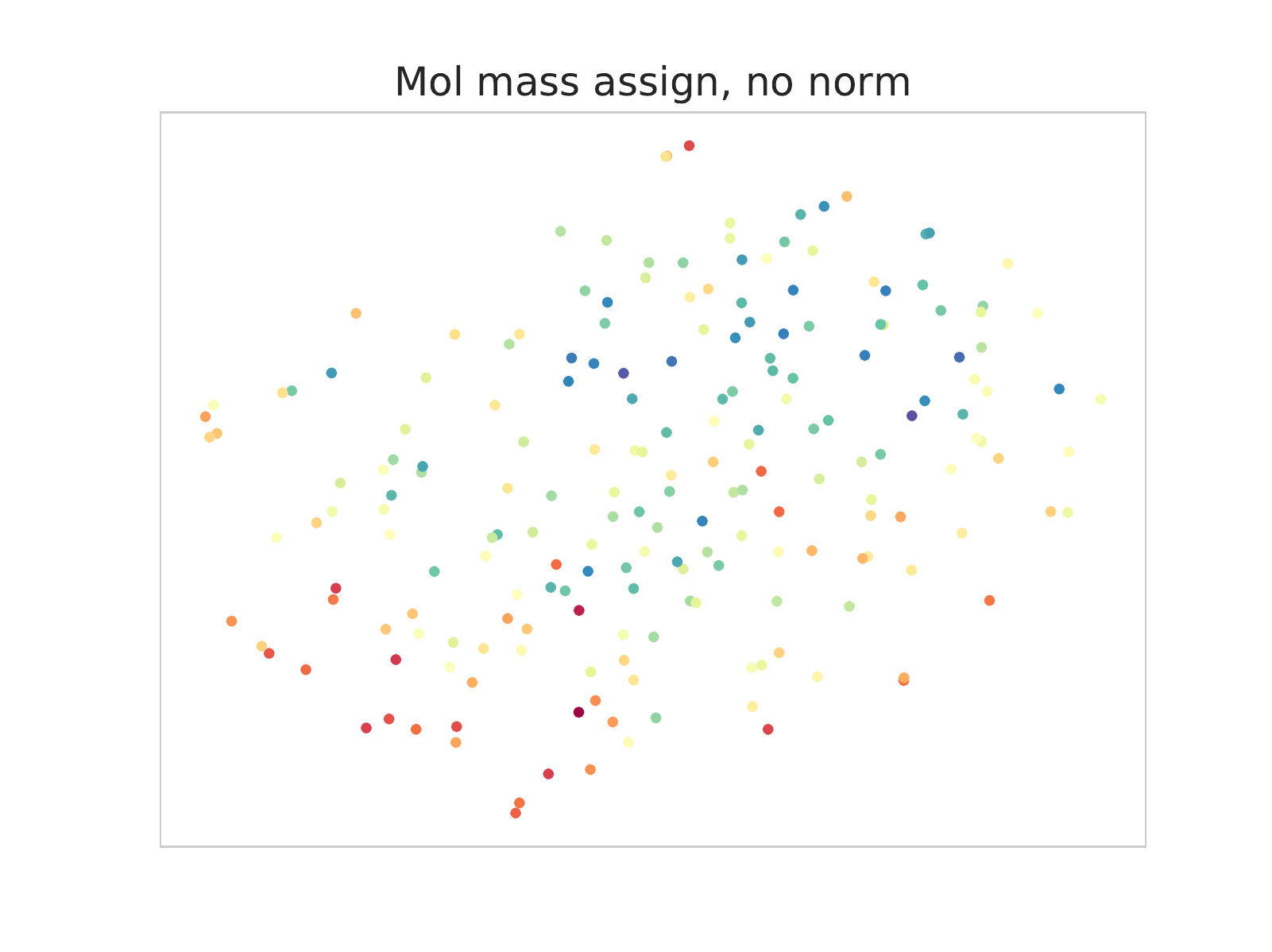}
&
\includegraphics[width=0.3\textwidth, clip=true, trim=6mm 0mm 6mm 0mm]{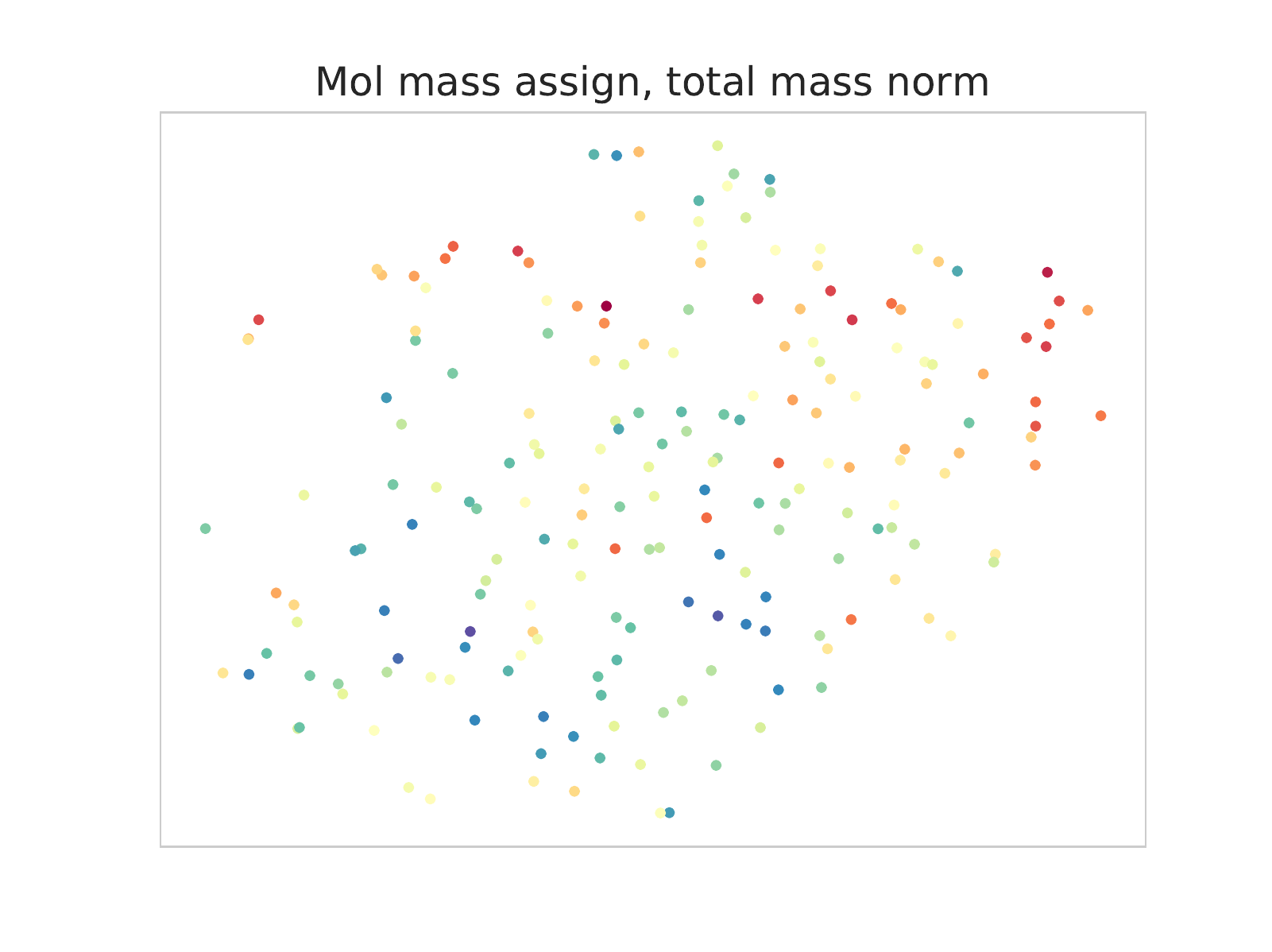}
\\[-1mm]
\includegraphics[width=0.3\textwidth, clip=true, trim=6mm 0mm 6mm 0mm]{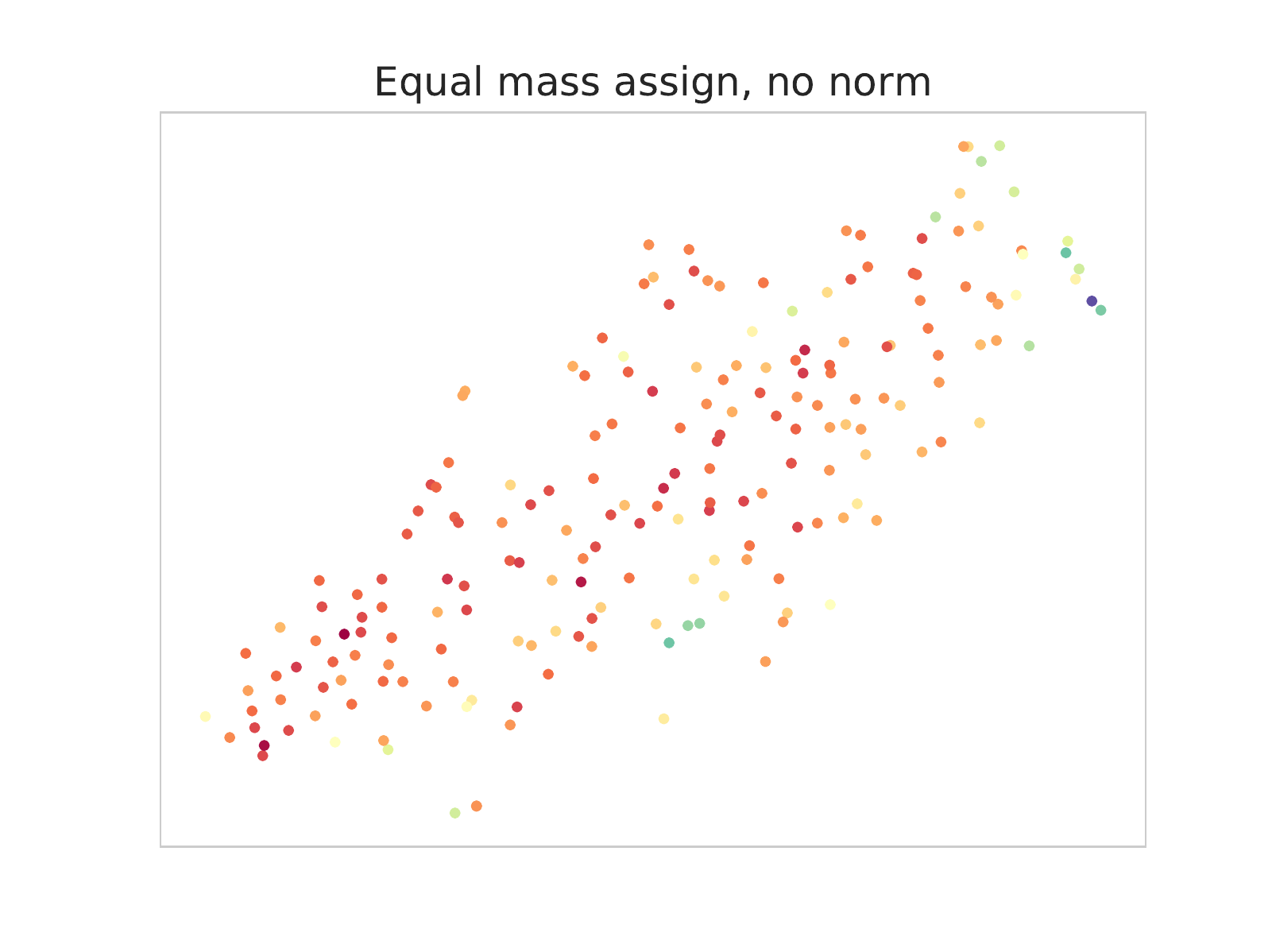}
&
\includegraphics[width=0.3\textwidth, clip=true, trim=6mm 0mm 6mm 0mm]{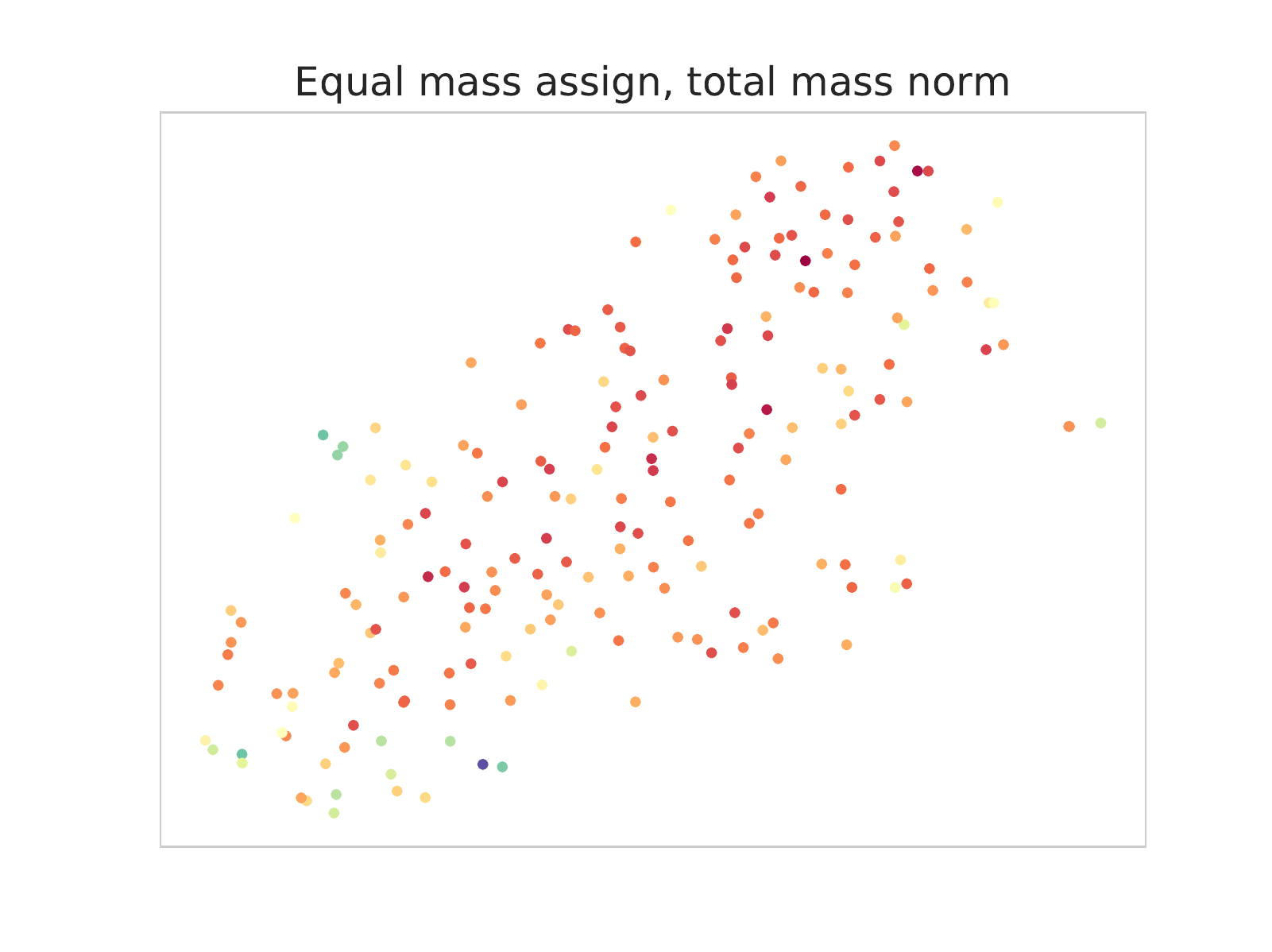}
\\
\includegraphics[width=0.3\textwidth, clip=true, trim=6mm 0mm 6mm 0mm]{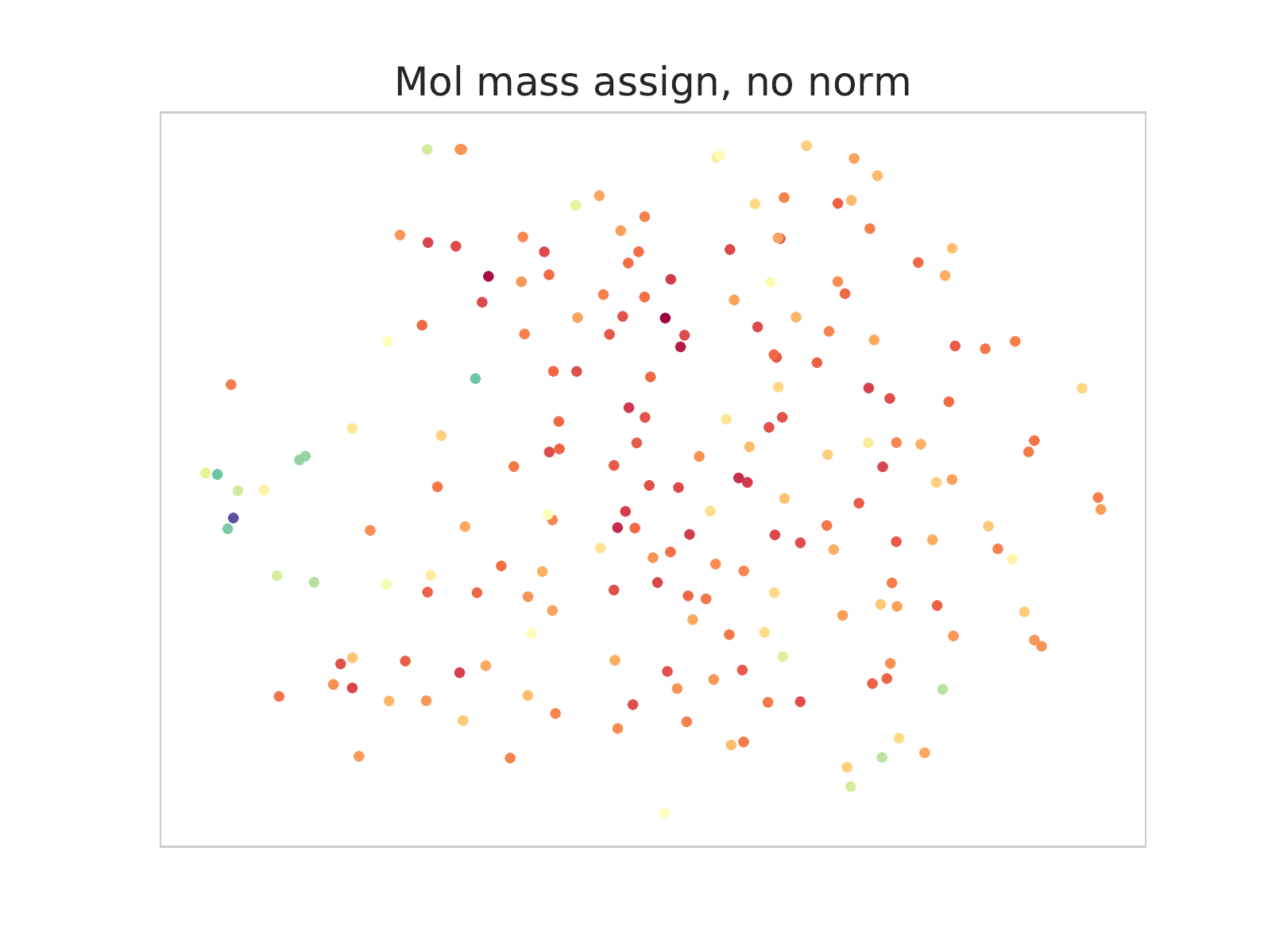}
&
\includegraphics[width=0.3\textwidth, clip=true, trim=6mm 0mm 6mm 0mm]{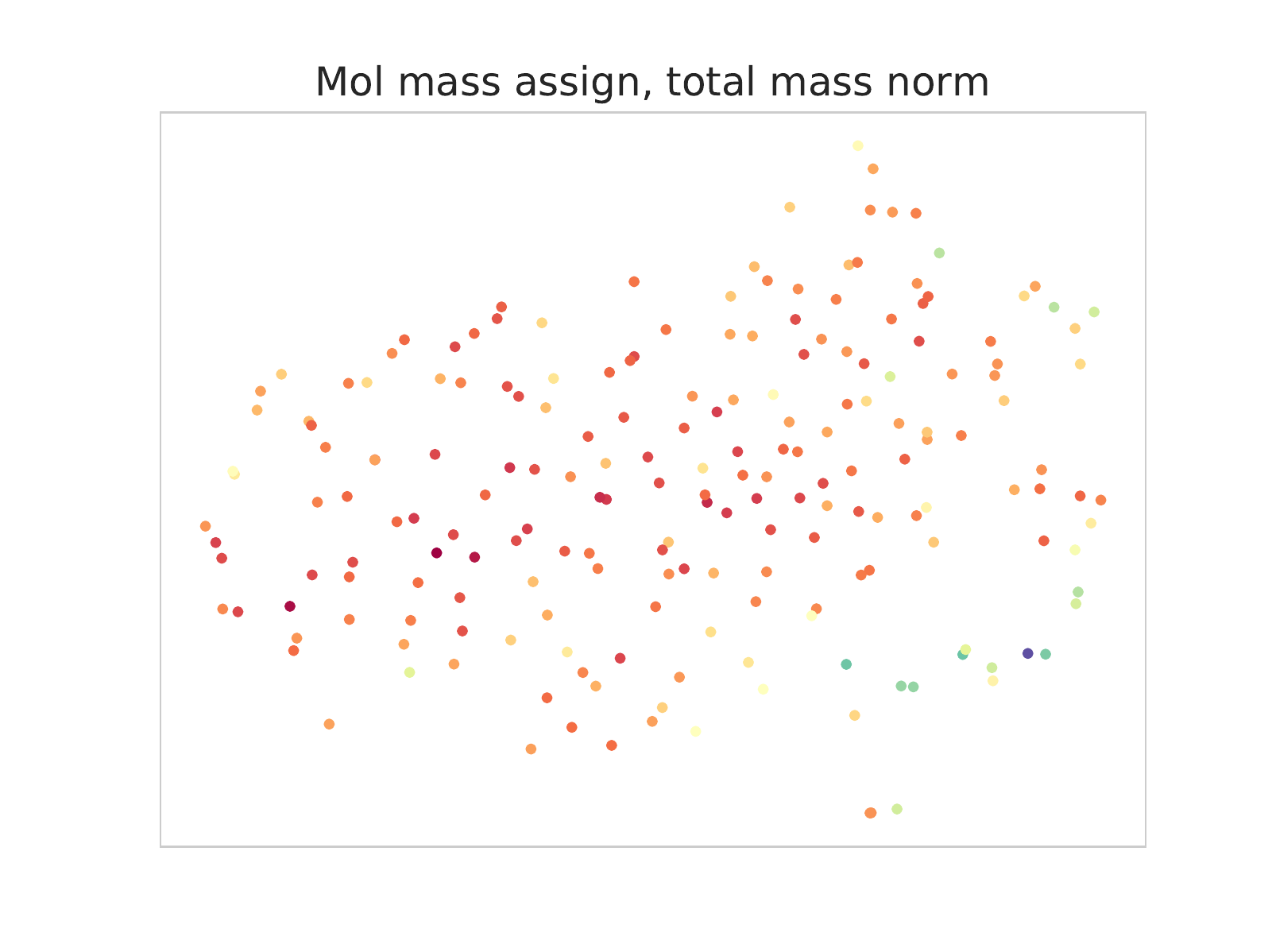}
\\[-1mm]
\end{tabular}
\caption{\small t-SNE visualization of OT distance \dist{} for different parameter configurations, first four color-coded by QED value, last four by SA score.}\label{fig:tsne_vis}
\end{figure*}
}

\newcommand{\insertPairwiseVis}{
\newcommand{\scatterimgwidth}{0.24\textwidth}
\begin{figure*}
\centering
 \includegraphics[width=\scatterimgwidth, clip=true, trim=4mm 0mm 1mm 0mm]{images/dist_vs_value_qed_1_copy.png}
 \includegraphics[width=\scatterimgwidth, clip=true, trim=4mm 0mm 1mm 0mm]{images/dist_vs_value_qed_2_copy.png}
 \includegraphics[width=\scatterimgwidth, clip=true, trim=4mm 0mm 1mm 0mm]{images/dist_vs_value_qed_3_copy.png}
 \includegraphics[width=\scatterimgwidth, clip=true, trim=4mm 0mm 1mm 0mm]{images/dist_vs_value_qed_4_copy.png}
\\
\includegraphics[width=\scatterimgwidth, clip=true, trim=4mm 0mm 1mm 0mm]{images/dist_vs_value_sascore_1_copy.png}
\includegraphics[width=\scatterimgwidth, clip=true, trim=4mm 0mm 1mm 0mm]{images/dist_vs_value_sascore_2_copy.png}
\includegraphics[width=\scatterimgwidth, clip=true, trim=4mm 0mm 1mm 0mm]{images/dist_vs_value_sascore_3_copy.png}
\includegraphics[width=\scatterimgwidth, clip=true, trim=4mm 0mm 1mm 0mm]{images/dist_vs_value_sascore_4_copy.png}
\captionsetup{font=small}
\caption{
\small Each point in the scatter plot indicates the dissimilarity measure
between the molecules (x axis)
and the difference in the QED score and SA score (y axis).
The four images are for the four different combinations of the distance. See text for interpretation.
\label{fig:pairwise}}
\end{figure*}
}

\newcommand{\insertPairwiseVisTwo}{
\begin{figure*}
\centering
 \includegraphics[width=.75\textwidth]{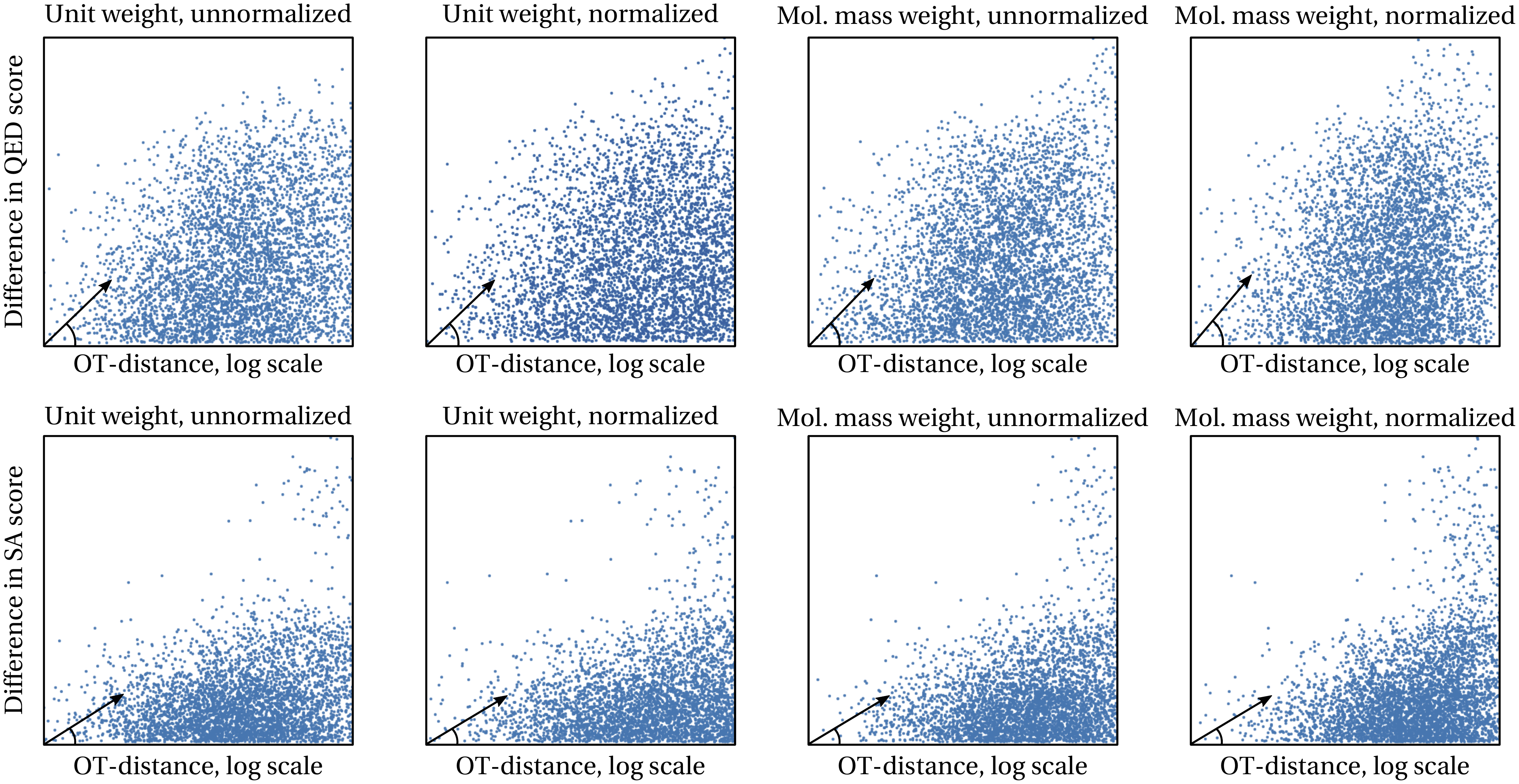}
\captionsetup{font=small}
\caption{
\small Each point in the scatter plot indicates the dissimilarity measure
between the molecules (x axis)
and the difference in the QED score and SA score (y axis).
The four images are for the four different combinations of the distance. See text for interpretation.
\label{fig:pairwise}}
\end{figure*}
}



\newcommand{\insertImprovementTable}{
\begin{table}[H]
\centering
\resizebox{\columnwidth}{!}{%
\begin{tabular}{|c|c|c|c|c|}
\hline
& \rand    & \fingerprint     & \dist  & \sumkernel    \\ \hline
QED      & $0.90 \pm 0.01$ & $0.91 \pm 0.01$ & ${0.93 \pm 0.01}$ & ${\bf 0.94 \pm 0.01}$ \\ \hline
P.logP & $6.81 \pm 0.34$ & ${\bf9.79 \pm 2.26}$ & $8.10 \pm 1.01$ & $8.65 \pm 0.43$ \\ \hline
\end{tabular}
}
\caption{\small Final value for QED and Pen-logP over 80 eval-s}
\label{tab:overall_res}
\end{table}
\vspace{-5mm}
}

\newcommand{\insertMoleculePics}{
\newcommand{\qedpicwidth}{0.13\textwidth}
\newcommand{\plogppicwidth}{0.21\textwidth}
\begin{figure*}[t]
\begin{center}
\begin{tabular}{@{}c@{\!\!\!\!\!}c@{\!\!\!\!\!}c@{\!\!\!\!\!}c@{}}
\includegraphics[width=\qedpicwidth, clip=true, trim=6mm 0mm 6mm 0mm]{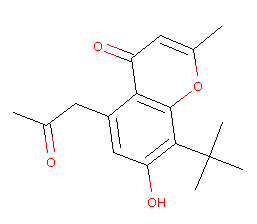}
&
\includegraphics[width=\qedpicwidth, clip=true, trim=6mm 0mm 6mm 0mm]{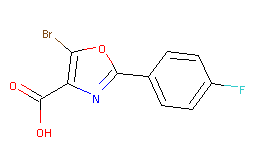}
&
\includegraphics[width=\qedpicwidth, clip=true, trim=6mm 0mm 6mm 0mm]{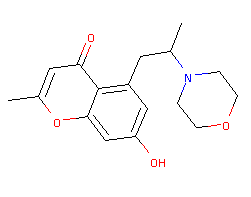}
&
\includegraphics[width=\qedpicwidth, clip=true, trim=6mm 0mm 6mm 0mm]{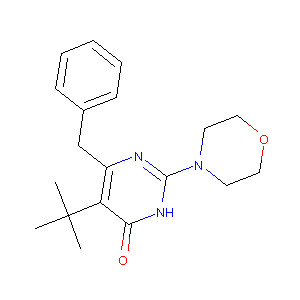}
\\[-2mm]
{\tiny (a) QED 0.92083} &{\tiny (b) QED 0.92145}  & {\tiny (c) QED 0.94023} & {\tiny (d)
QED 0.94087}  \\
\includegraphics[width=\plogppicwidth, clip=true, trim=0mm 2mm 6mm 7mm]{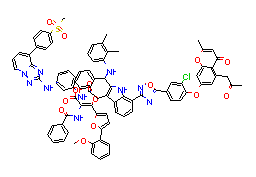}
&
\includegraphics[width=\plogppicwidth, clip=true, trim=0mm 2mm 6mm 7mm]{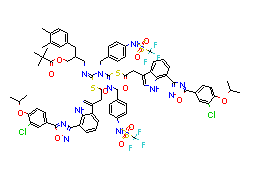}
&
\includegraphics[width=\plogppicwidth, clip=true, trim=0mm 2mm 6mm 7mm]{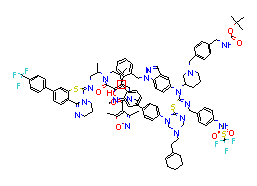}
&
\includegraphics[width=\plogppicwidth, clip=true, trim=0mm 2mm 6mm 7mm]{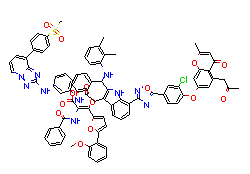}
\\[-2mm]
{\tiny (e) plogp 11.271} & {\tiny (f) plogp 11.988} & {\tiny (g) plogp 12.231} & {\tiny (h) plogp 11.270} \\
\end{tabular}
\end{center}
\captionsetup{font=small}
\caption{\small A random sample of optimal molecules and values found by \Chem{}.
In the top row, we show those with the highest QED scores, and in the bottom row we show the same for Pen-logP.
\label{fig:molecule_pics}
}
\end{figure*}
}

\newcommand{\insertOptimGraphs}{
\begin{figure*}
\centering
\includegraphics[width=2.7in]{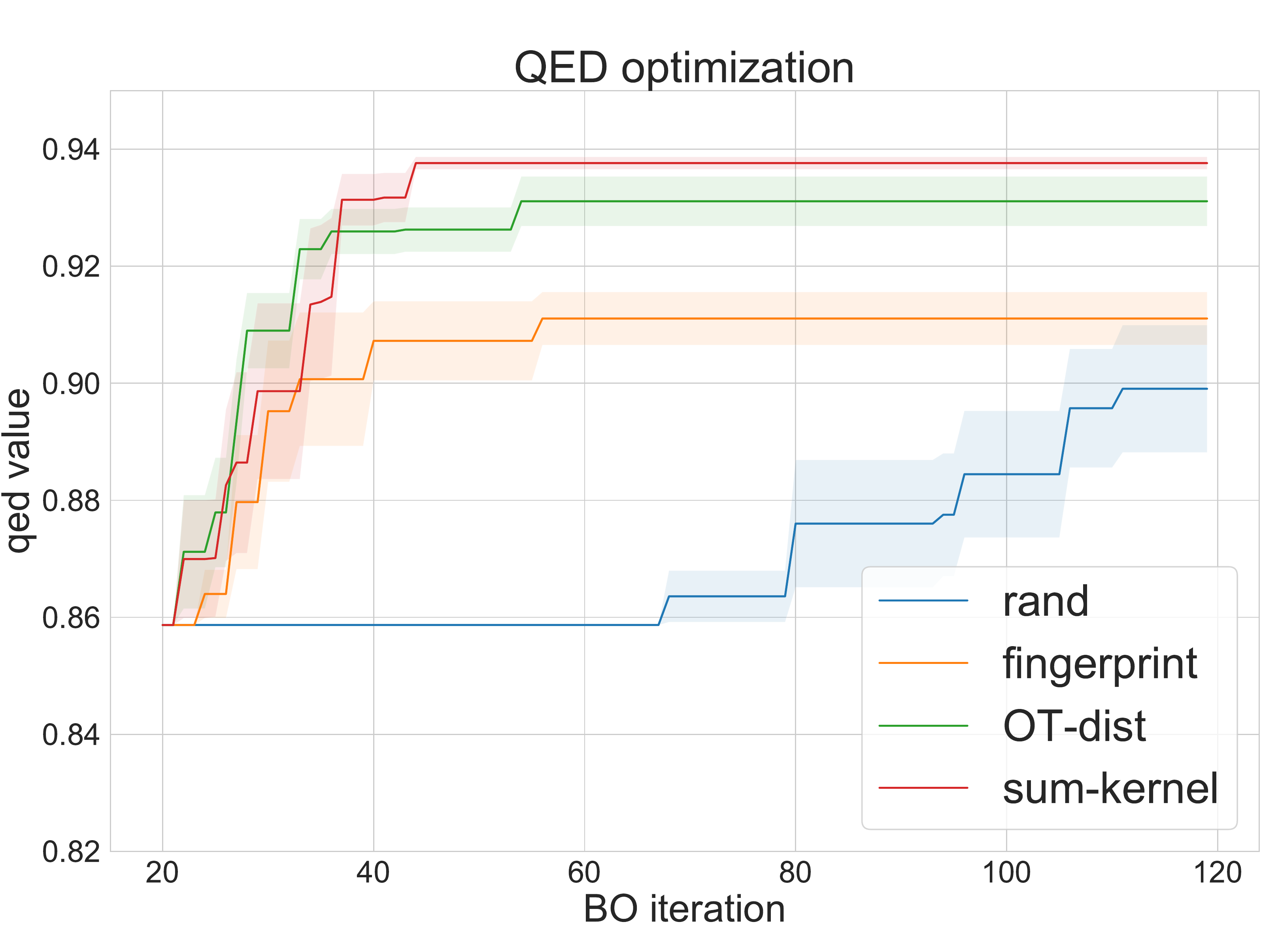}
\includegraphics[width=2.7in]{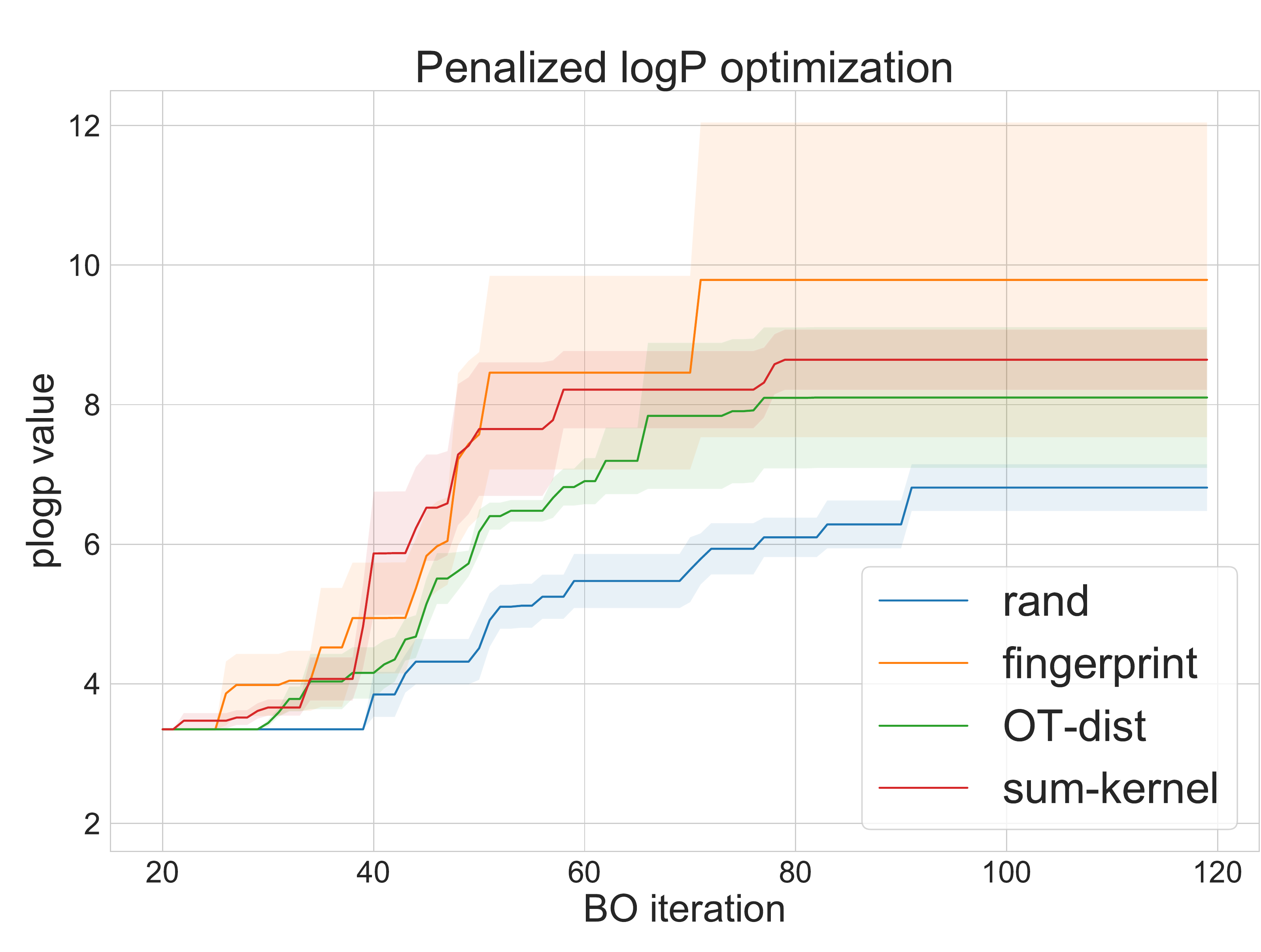}
\caption{\small
Results comparing the three methods described in the beginning of
Section~\ref{sec:experiments}.
We plot the number of iterations (after initialization) against the highest
found QED (left) and Pen-LogP (right) values by each
method, where higher is better.
All curves were produced by averaging over $5$ independent runs.
The shaded regions indicate one standard error.
\label{fig:optim_graphs}
}
\end{figure*}
}

\newcommand{\insertSynPaths}{
\begin{figure*}
\caption{\small
Synthesis path for molecule with penalized logP $11.988$.
The boxed molecules are from the initial pool of $20$ reagents.
\label{fig:syn_path1}
}
\centering
\includegraphics[width=5in]{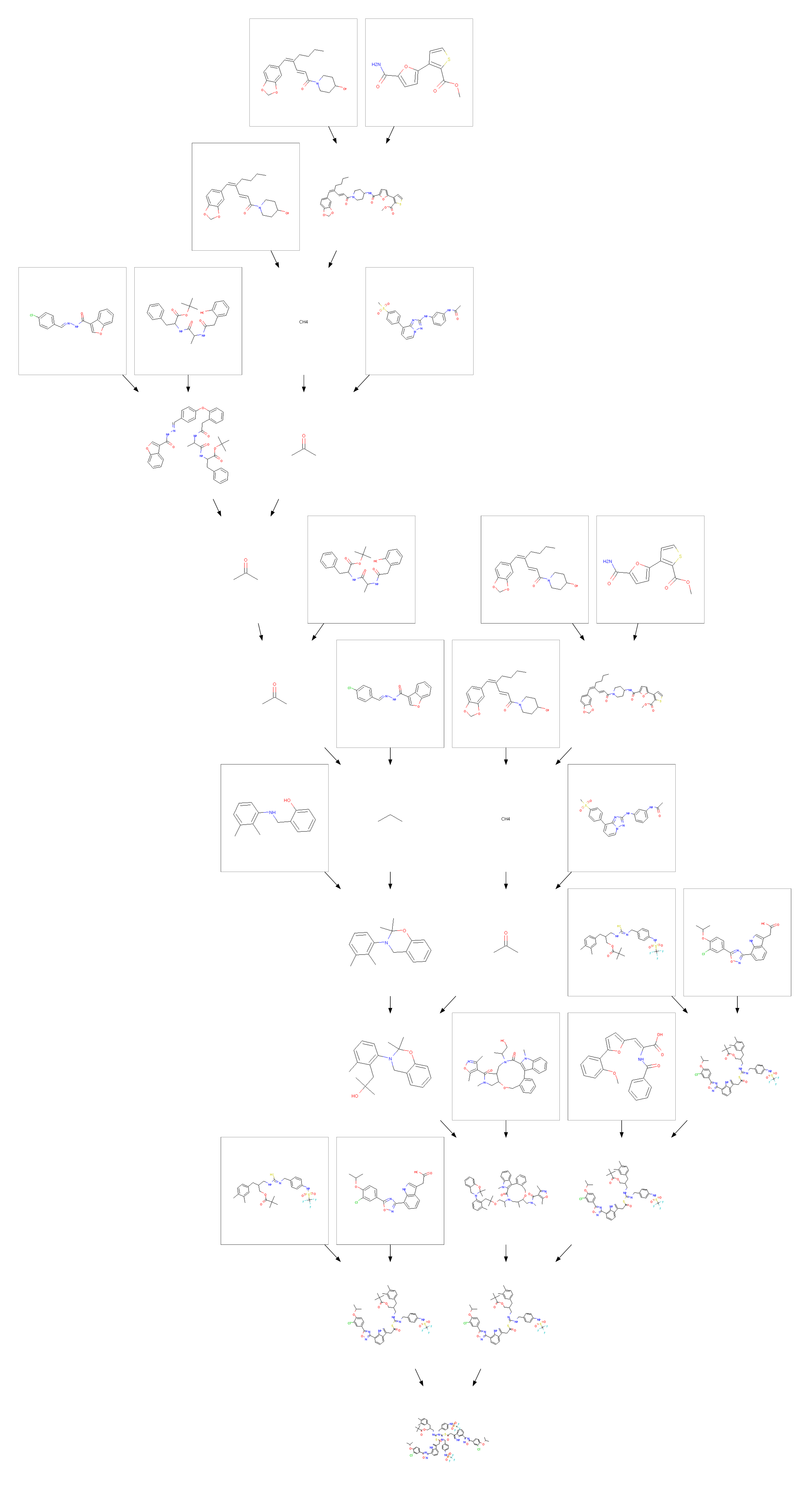}
\end{figure*}

\begin{figure*}
\caption{\small
Synthesis path for molecule with QED $0.92$.
The boxed molecules are from the initial pool of $20$ reagents.
\label{fig:syn_path2}
}
\centering
\includegraphics[width=5in]{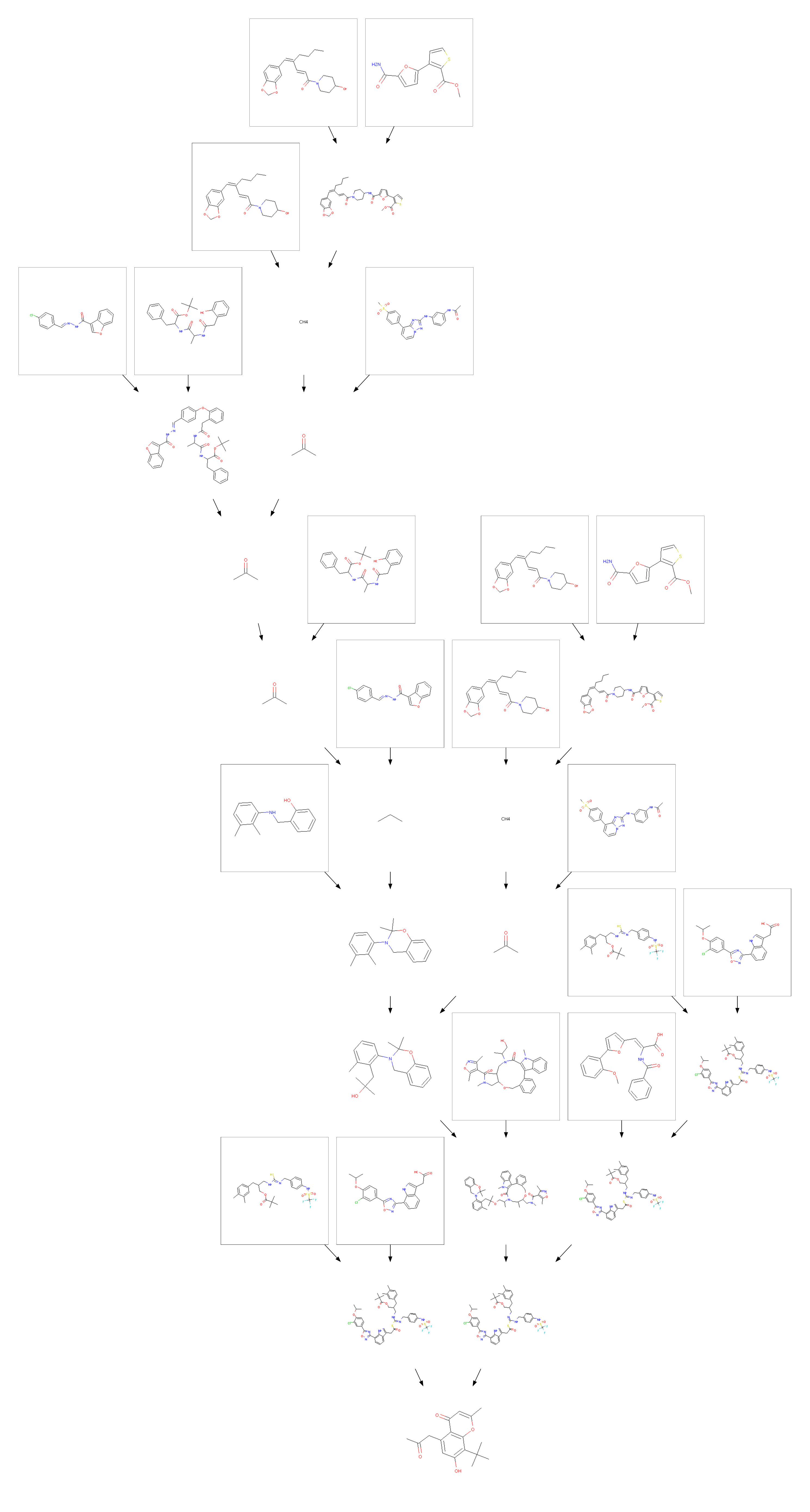}
\end{figure*}

\begin{figure*}
\caption{\small
Synthesis path for molecule with penalized logP $8.306$.
The boxed molecules are from the initial pool of $20$ reagents.
\label{fig:syn_path3}
}
\centering
\includegraphics[width=5in]{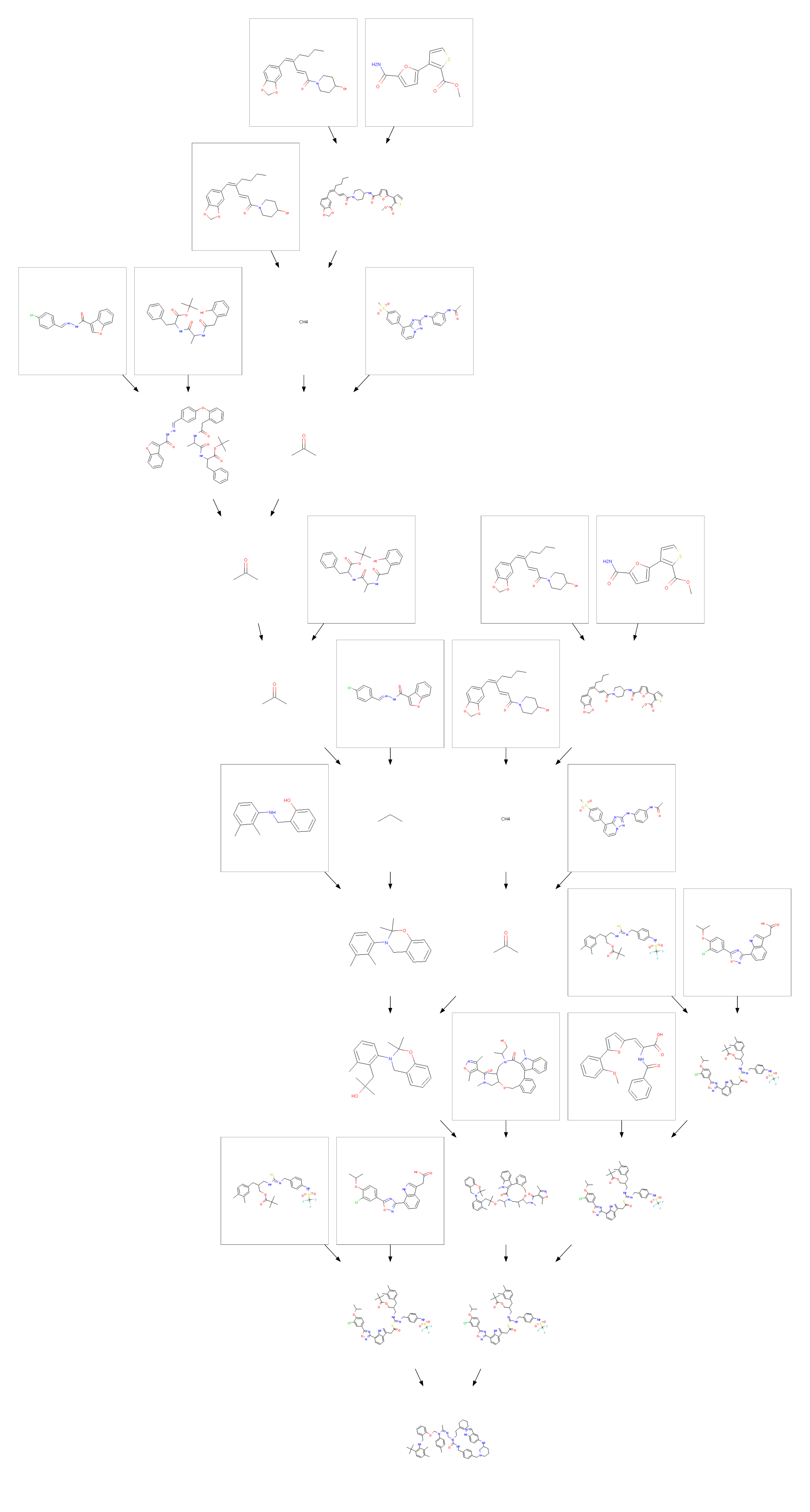}
\end{figure*}

\begin{figure*}
\caption{\small
Synthesis path for molecule with QED $0.93$.
The boxed molecules are from the initial pool of $20$ reagents.
\label{fig:syn_path4}
}
\centering
\includegraphics[width=5in]{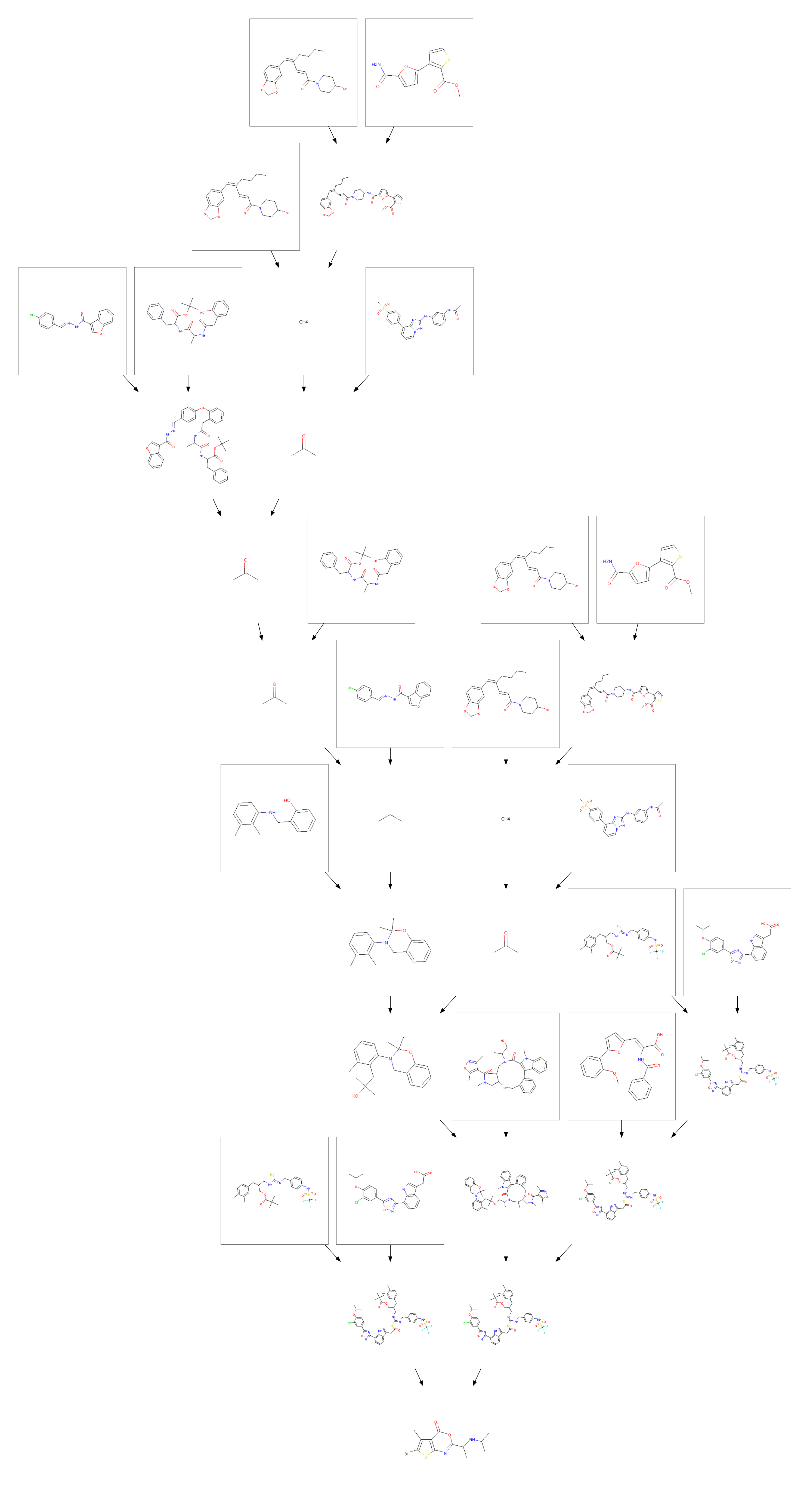}
\end{figure*}
}

\newcommand{\insertOptimGraphsLowStart}{
\begin{figure*}
\centering
\includegraphics[width=2.3in]{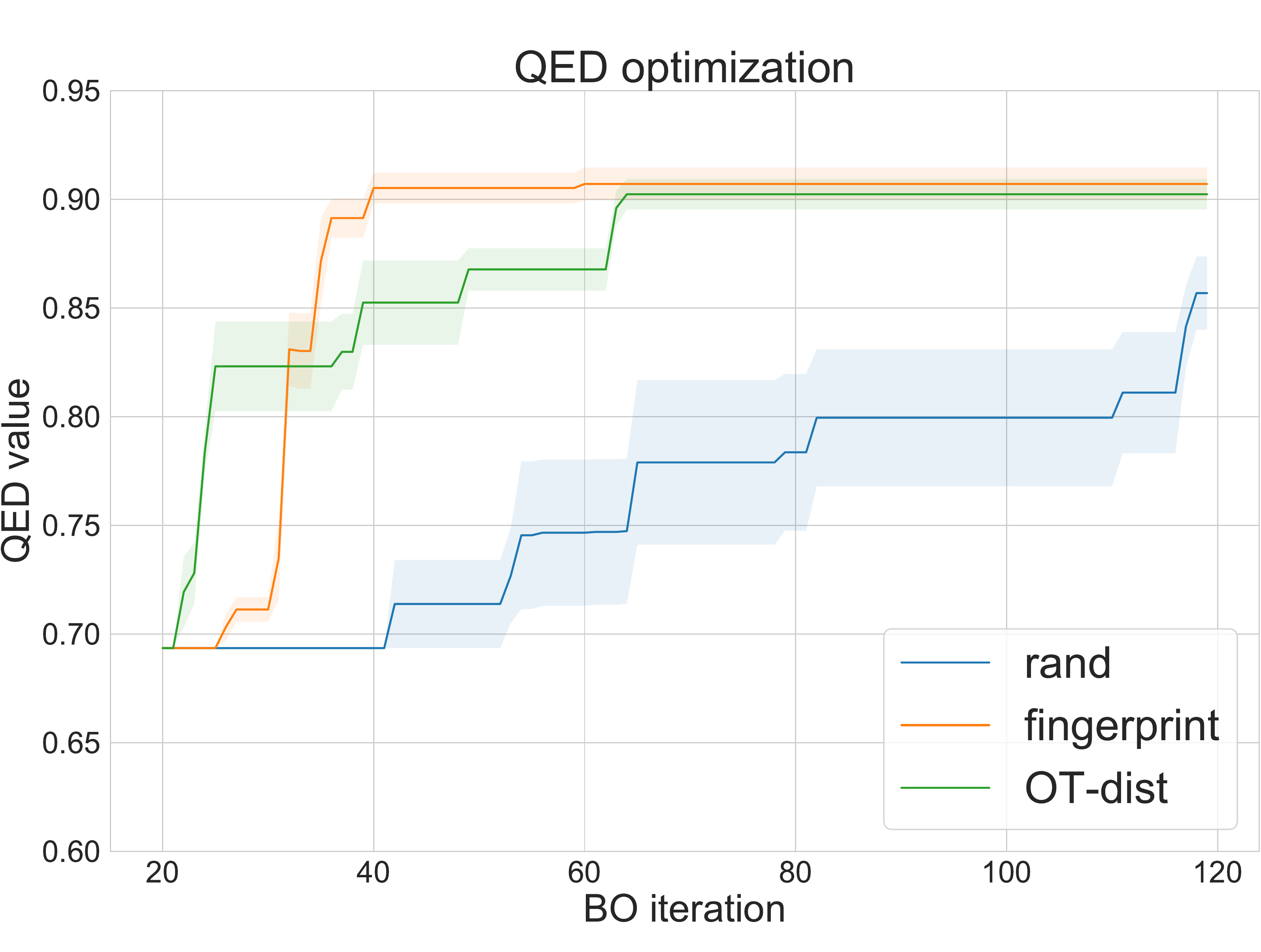}
\hspace{0.1in}
\includegraphics[width=2.3in]{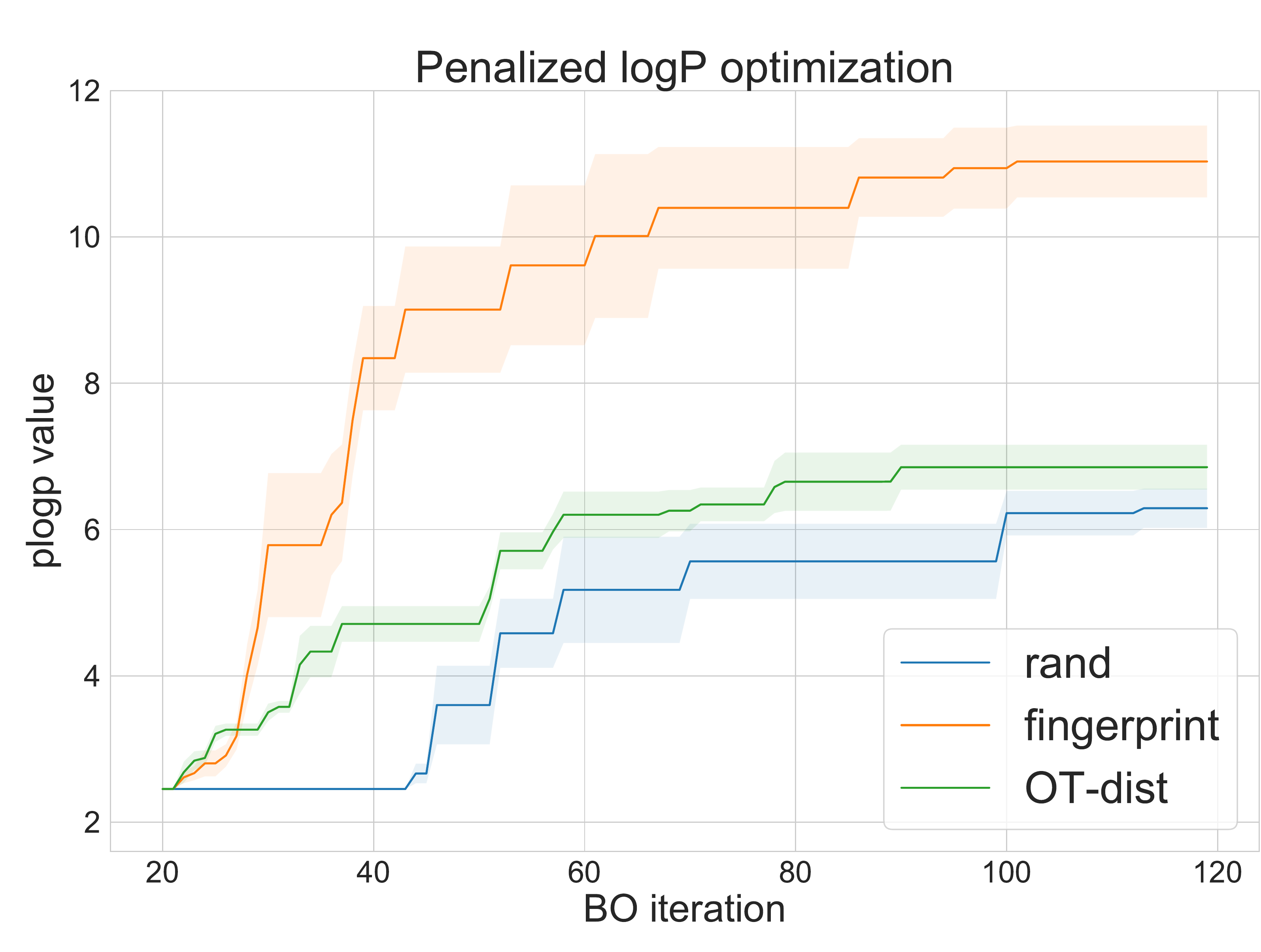}
\caption{\small
Results comparing the three methods described in the beginning of
Section~\ref{sec:experiments}.
We plot the number of iterations (after initialization) against the highest
found QED (left) and Pen-LogP (right) values by each method.
Higher is better in both cases.
All curves were produced by averaging over $5$ independent runs.
The shaded regions indicate one standard error.
\label{fig:optim_graphs_low}
}
\end{figure*}
}

\newcommand{\insertTsneVisDifferent}{
\begin{figure*}
\includegraphics[width=0.3\linewidth, clip=true, trim=6mm 0mm 6mm 0mm]{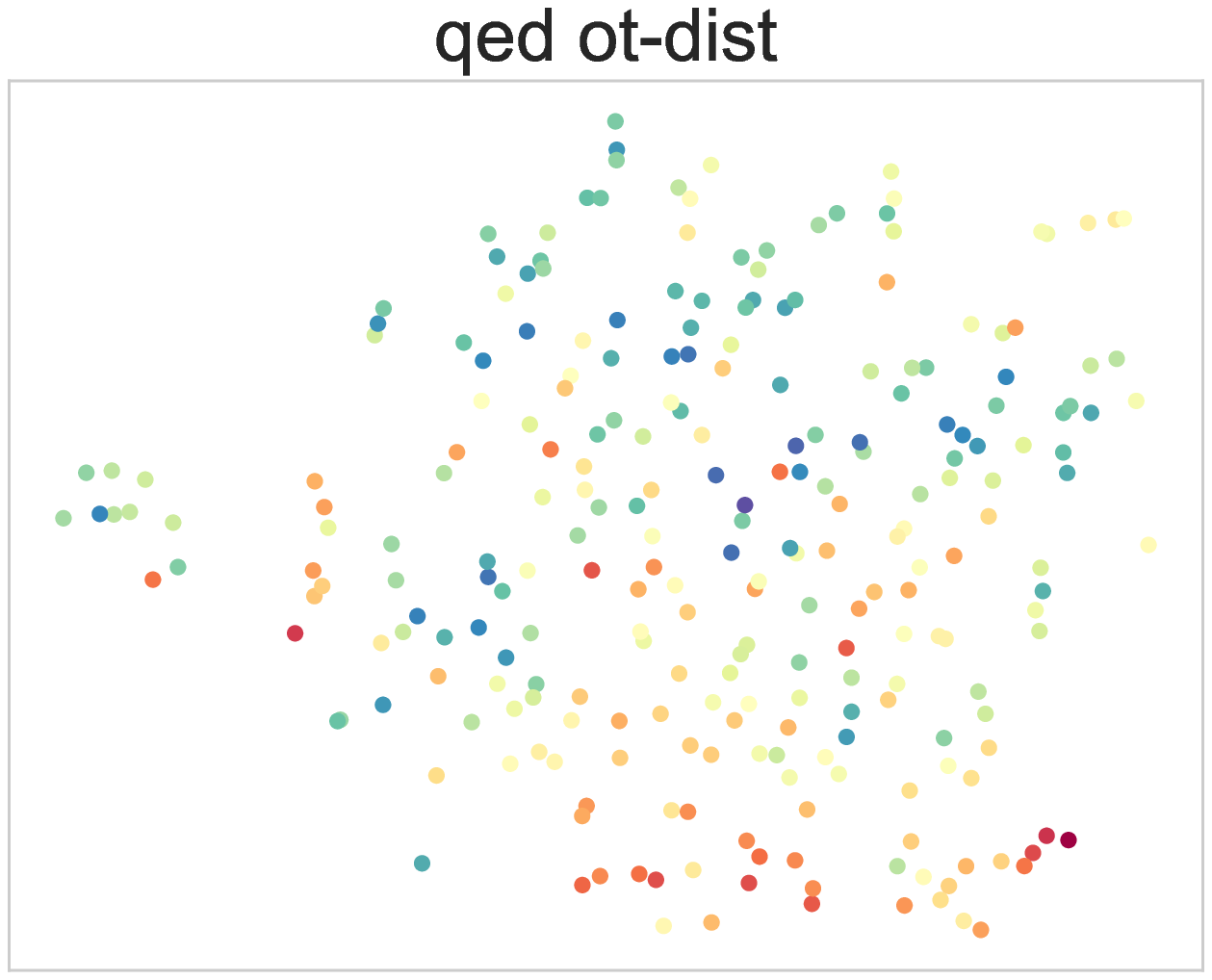}
\includegraphics[width=0.3\linewidth, clip=true, trim=6mm 0mm 6mm 0mm]{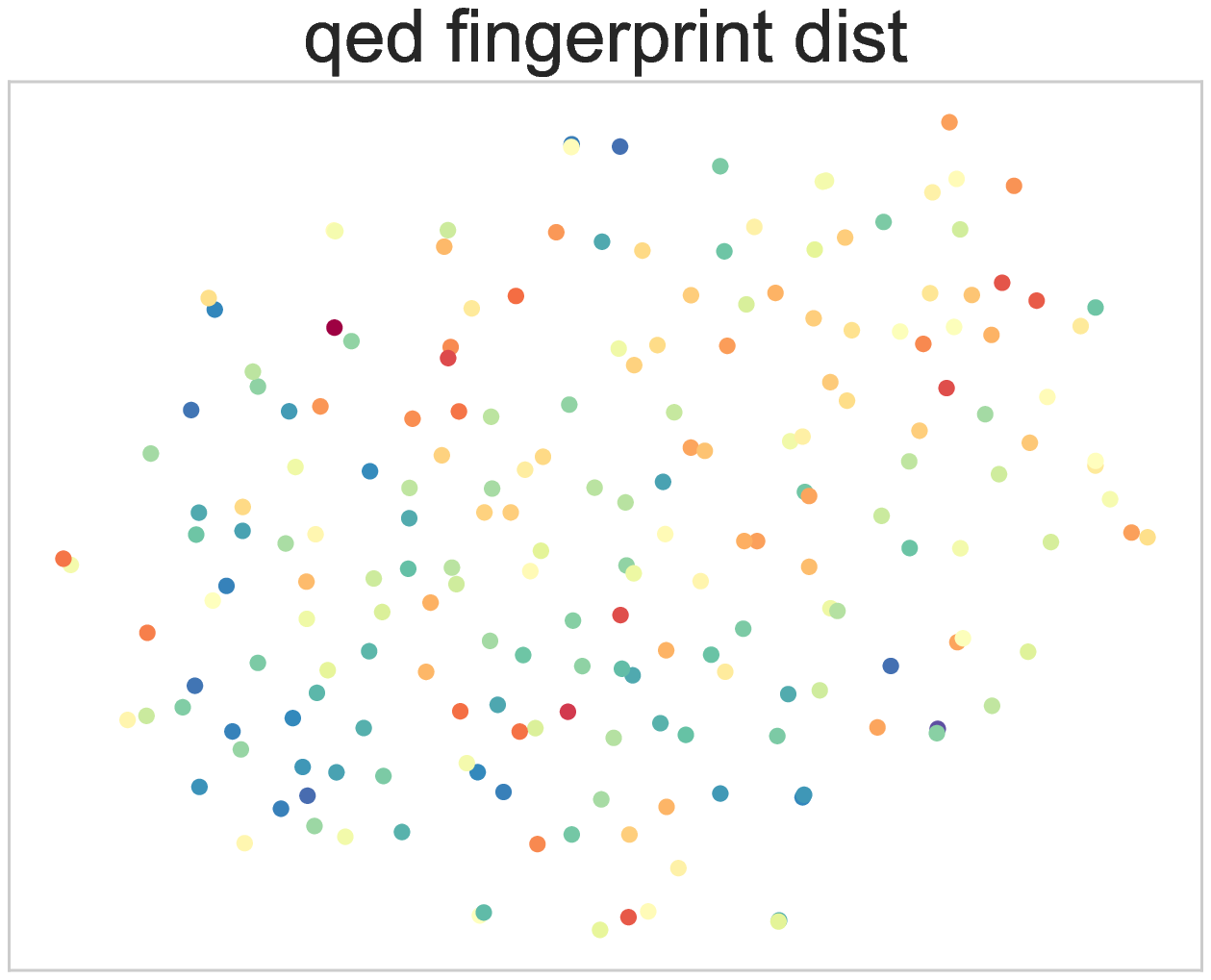}
\includegraphics[width=0.3\linewidth, clip=true, trim=6mm 0mm 6mm 0mm]{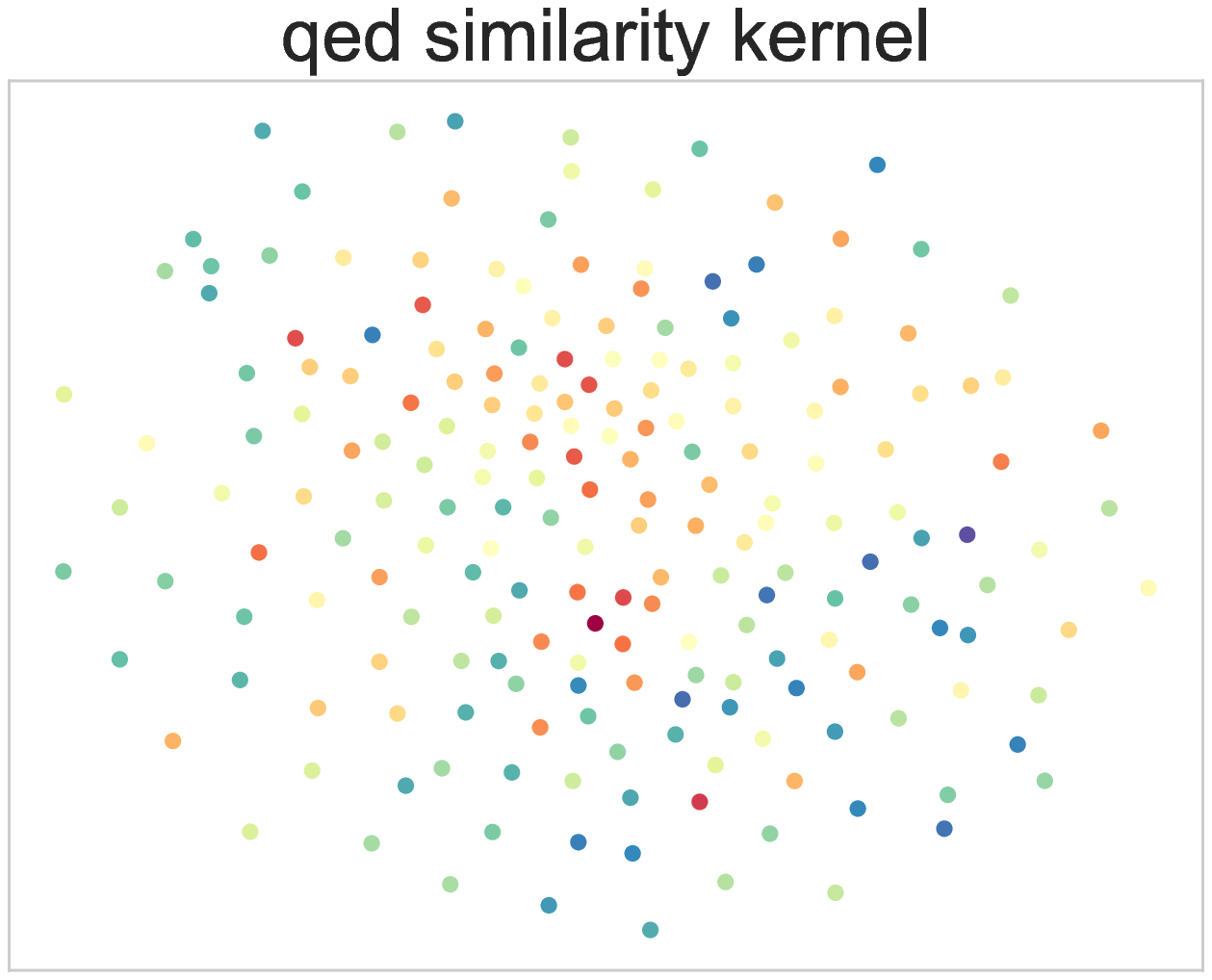}
\caption{\small
Comparison of t-SNE embeddings produced based on three molecular distances: \dist{}, $\ell_2$ distance between fingerprint vectors, and inverted similarity kernel between fingerprints.
\label{fig:tsne_vis2}
}
\end{figure*}
}

%% file: abstract.tex
\begin{abstract}
In applications such as molecule design or drug discovery, it is desirable to have an algorithm which recommends new candidate molecules based on the results of past tests. These molecules 
first need to be synthesized and then tested for objective properties. We describe \Chem, a Bayesian optimization framework for generating and optimizing organic molecules for desired molecular properties. While most existing data-driven methods for this problem do not account for sample efficiency or fail to
enforce realistic constraints on synthesizability, our approach explores the synthesis graph in a sample-efficient way and produces synthesizable candidates. We implement \Chem{} as a Gaussian process model and explore existing molecular kernels for it. Moreover, we propose a novel optimal-transport based distance and kernel that accounts for graphical information explicitly. In our experiments, we demonstrate the efficacy of the proposed approach on several molecular optimization problems.

\end{abstract}

%% file: introduction.tex
\section{Introduction}
\label{sec:intro}


In many applications, such as drug discovery and materials optimization, one is interested in
designing chemical molecules with desirable properties~\citep{dl_review}.
For instance, in drug discovery, one wishes to find molecules with high solubility in
blood and high potency, but low toxicity.
Recently, we have seen a surge of interest in the adoption of machine learning techniques
for such tasks, due to their effectiveness in modeling structure-property relations
of molecules, and due to limitations of traditional computational chemistry methods in
effectively exploring the large and complex space of chemical molecules.
For instance, the number of drug-like molecules is estimated to be between
$10^{23}$ and $10^{60}$~\citep{estim_druglike}, among which only around $10^8$ have been
synthesized.
While there have been several strategies for this problem, such as generative modeling,
reinforcement learning, and more~\citep{bomb,jt-vae,release,gcpn,oliynyk2016high}, one
promising approach is to treat this task as a black-box optimization problem (e.g.
~\cite{ling2017high,constr_bo}).
Here, we  assume the existence of a function $f:\Xcal\rightarrow \RR$ defined on the
chemical space $\Xcal$, where $f(x)$  is a measure of goodness of molecule $x$ for the
relevant application.
The goal is to find the optimum of this function
$\argmax_{x\in\Xcal}f(x)$.
In real world settings, $f$ is typically derived from the results of laboratory
experiments.
The algorithm would then use results of the past experiments, i.e the $f(x)$ values, to
recommend new molecules.
Since conducting such experiments are expensive, it is
imperative to find the maximum in as few evaluations as possible.

In this work, we contribute to this line of research by
developing \Chem, a Bayesian optimization (BO) framework for generating and optimizing
molecules, focusing on small\footnote{%
In contrast with biologics (large molecules), which are protein based.
} organic molecules for drug discovery.
In doing so, we wish to emulate a real world setting,
where an algorithm would recommend new candidate molecules.
These molecules \emph{first
need to be synthesized}, and then tested for necessary
properties. Ideally, the algorithm would not only ensure that the recommended molecule is
chemically valid and synthesizable, but also provide a recipe for synthesis and take into consideration the reagents and resources available.
Even in cases where the recommended molecules are synthesized manually,
providing a recipe can be a helpful guide to
the chemist and greatly reduce the amount of manual work required. Combining sequential decision making and synthesis, \Chem{} is a first step towards automated molecular optimization. To summarize, our contributions are:
\begin{enumerate}[leftmargin=0.6cm]
\item 
We develop a Gaussian process (GP) model of structure-property relations in molecules.
For the GP kernel, we use prior work on molecular
fingerprints~\citep{ralaivola2005graph,hinselmann2010graph}
and additionally design a new optimal transport based similarity measure between
molecules by treating them as graphs.
\item
We use a synthesis graph to navigate the chemical space.
On each iteration of BO, \Chem{} recommends the molecule on this synthesis graph
that is deemed to be the most promising by the GP,
i.e. the molecule with
the highest \emph{acquisition} value~\citep{brochu2010tutorial}.
This approach not only ensures that each recommended molecule is chemically valid,
but also provides a synthesis recipe\footnote{%
We will qualify this statement later in Section~\ref{sec:explorer}.
}.
\item In our experiments, we demonstrate that \Chem{} outperforms simpler alternatives
for synthesizeable optimization, which do not use a probabilistic model to
guide search.
The final values for the popular QED~\citep{qed} and
penalized partition coefficient~\citep{constr_bo} benchmarks
achieved by \Chem{} are competitive with state-of-the-art methods, while
using significantly less data and function evaluations.
Our code is released open source at \url{https://github.com/ks-korovina/chembo}.
\end{enumerate}

\section{Related Work}\label{related_work}

\paragraph{Optimization:}
SMILES strings~\citep{smiles}, which describe the structure of molecules as a
string, are a common representation used in machine learning techniques for molecular
optimization~\citep{bomb,cho18}.
One of the main reasons for their adoption is that SMILES strings allow one to use existing NLP machinery largely unchanged.
Recently, graph representations for molecules have become popular.
Most recent methods adopting this representation use generative models or reinforcement learning to construct a molecular graph, and optimize the property in question while attempting to maintain validity~\citep{jt-vae,organ,gcpn,zhou2018optimization,multimodal}. In learning representations for molecules, they draw on the methods that process graph data directly, such as graph neural networks \citep{gcpn, multimodal} and covariant compositional networks  \citep{hy2018predicting}.
However, drug/materials optimization is a \emph{stateless} optimization
problem, where there is no explicit need to deal with states and solve credit assignment.
This can require a large number of samples~\citep{jiang2017contextual}, and is not desirable
 in settings where each evaluation might involve several
laboratory experiments.
BO methods, which are particularly well suited for optimization problems with expensive
evaluations, are sparsely represented in the field.%
~\citet{bomb, grammar_vae, jt-vae} learn a
Euclidean representation for molecules and perform BO on this space,
while~\citet{constr_bo} extend that work to account for validity constraints.

\paragraph{Synthesizable recommendations:}
In much of the above work, synthesizability of recommendations remains one of the most
important concerns.
The common approach to tackle this problem is to consider a proxy synthesizability
score, by either imposing search constraints on the objective~\citep{constr_bo} or
incorporating the score into $f$ along with the other
properties~\citep{bomb}.
However, synthesizability scores are not always reliable.
For example,~\citet{bomb} found that their autoencoder produced a large number of
molecules with unrealistically large carbon rings when using the SA synthesizability
score~\citep{sascore} as the reward function.
More critically, it ignores practical challenges in a laboratory environment.
First, a chemist may not have the reagents and/or the process conditions
available to synthesize the molecule.
Second, figuring out the synthesis plan for a single molecule,
let alone the hundreds of them recommended during the optimization routine, 
can be quite challenging.
While one could consider using retrosynthesis techniques~\citep{law2009route} for
the latter, they may not always be reliable, and moreover, one can run into the same
availability problems mentioned above.

We leverage a large and separate direction of research which use
ML techniques to predict outcomes of chemical reactions%
~\citep{schwaller2018molecular,nnpred16,coley-rexgen19,%
chen2009no,law2009route}
(see~\citet{engkvist2018computational} for a more complete list).
The first methods for such
\emph{synthesis prediction} tasks were template based,
in that they either select relevant rules from a fixed library, or rank enumerated
outcomes of applying these rules.
One of the first examples in the ML community was~\citet{nnpred16}, which
predicted reaction type and then used SMARTS transformations to construct
candidate outcome graphs.
Due to rigidity of template-based approaches, template-free
methods have become increasingly popular~\citep{wldn17,schwaller2018found}.
One such method in this class, and the one we adopt in this work to explore the
chemical space, is Rexgen~\citep{coley-rexgen19}.
It proceeds in two stages: first, reactive sites are predicted using a
Weisfeiler-Lehman network with global attention~\citep{wln17} on the graph
representation of
reaction inputs; next, possible configurations of connectivity changes in reactive sites
are scored with a Weisfeiler-Lehman Difference network~\citep{wldn17}.

\paragraph{Joint optimization and synthesis:} Our approach in this paper, which marries both
directions of work, can be viewed in two ways.  On one hand, it performs optimization
while ensuring the recommendations are synthesizable. On the other hand, as we will
explain shortly, it explores synthesis paths to discover promising candidate molecules via
a data-driven guide. This approach is the core novelty of our work. As far as we are
aware, there is only one work in this direction: concurrently with us,
\citet{bradshaw2019model} pursued a similar goal of performing optimization with synthesis
guarantees.  However, their methodology and outcomes are very different from ours, in that
they adopt a generative model on subsets of molecules and train it jointly with a property predictor directly on that latent space, all of which may require many
samples.
As a result, while their method produces useful representations for
subsets of molecules,
unlike \Chem{}, it is not designed for sample-efficient goal-directed optimization tasks.

\paragraph{Kernels on molecules:}
For our GP based BO approach, we need to define a kernel between molecules.
While there has been prior work on defining kernels and similarity 
metrics between
graphs~\citep{wallis2001graph,kondor2002diffusion,sutherland2015scalable},
most do not account for more complex
properties of molecules in addition to graphical structure.
There have been a variety of neural network based graph similarity measures proposed for
molecules~\citep{liu2018chemi,torng2018graph,kearnes2016molecular}.
However, these approaches are computationally
expensive, which can be challenging in our GP based approach, where the similarity needs to be
computed for several pairs of molecules during each iteration of the BO routine.
A common class of graph based kernels used in chemoinformatics are based on
molecular fingerprints~\citep{ralaivola2005graph,hinselmann2010graph},
which have been found to outperform conventional graph kernels on some
tasks~\cite{sun2014}.
In \Chem{}, we use one such molecular fingerprint kernel in our GP.
However, molecular fingerprints essentially featurize the graph attributes and might
not  capture all necessary graphical information.
For this reason,
we develop a novel graph based similarity measure between molecules which is
computed via an optimal transport program.
It is most similar to~\citet{kandasamy2018nasbot} who use an
optimal transport based kernel for neural architecture search.
In our experiments, we found that while the performance of the molecular fingerprint
kernel and our dissimilarity measure can depend on the objective, they generally
outperform naive strategies which do not use a probabilistic model to inform
recommendations.

%% file: method.tex
\section{Method}
\label{sec:method}


\subsection{\Chem{} as a Gaussian Process based Bayesian Optimization Algorithm}

In this work, we design a Gaussian process (GP) based Bayesian optimization (BO) procedure. The reader can find a detailed review of GP-based BO in ~\citep{snoek12practicalBO,dragonfly}, and specific implementation details (such as the choice of acquisition function) in Appendix~\ref{sec:model_details}. Here we focus on the two central decisions of designing a GP based BO solution for molecular optimization: \textit{choosing a GP kernel} to specify a GP model, and \textit{designing a method to optimize the acquisition function}. In Section~\ref{sec:kernel}, we specify GP models,
specifically choices for the kernel $\kernel(x,x')$ between
two molecules $x$ and $x'$.
Next, in Section~\ref{sec:explorer}, we describe a method to optimize the acquisition
$\varphi_t$ over the chemical space $\Xcal$.
As mentioned previously, when doing so, we will strive to ensure that 
the recommendations are synthesizable and provide a synthesis recipe.
We note that while there are several options for
the kernel and the acquisition optimization
strategy in conventional domains, such as
Euclidean spaces, both tasks are nontrivial
in the chemical space
and constitute the major contributions of this work. We outline 
the \Chem{} procedure in Algorithm~\ref{algo:chembo}.

\begin{algorithm}[t]
\small
\caption{\Chem{}}
\label{algo:chembo}
\begin{algorithmic}[1]
  \State \textbf{Input:} Number of steps $T$, Initial evaluations $D_0$
  \For{$t=1, \ldots, T$}
    \State Infer posterior $\text{GP}(\mu_t(x), \kappa_t(x,x') | D_{t-1})$
    \State $x_t \leftarrow \argmax_{x \in \mathcal{X}} \varphi_t(x)$ \hspace{4mm} $\triangleright$ \textproc{Acquisition-Opt}
    \State $f(x_t) \leftarrow \textproc{Evaluate } x_t$
    \State $D_t \leftarrow D_{t-1} \cup \{x_t, f(x_t)\}$
  \EndFor
  \Return $x^* \leftarrow
  \argmax_{x_t \in \{x_1, \ldots, x_T \} } f(x_t)$
\end{algorithmic}
\end{algorithm}

\subsection{Kernel}
\label{sec:kernel}

A natural option would be to simply use one of the existing molecular kernels.
Indeed, molecular fingerprint based kernels are known to work well for several
applications, and we use that of~\citet{ralaivola2005graph} in \Chem.
However, they may not be able to capture all graphical information, which motivates
us to develop a new similarity measure described below.

\textbf{An optimal transport based kernel:}
We will describe a dissimilarity measure $\otdistsymbol:\Xcal^2\rightarrow\RR_+$ between
molecules. Given such a measure, $\kernel = e^{-\beta \otdistsymbol}$ where $\beta>0$,
is a similarity measure which can be used as a kernel.
The graphical structure of a molecule determines many of its chemical properties,
and as such, our measure will view molecules as graphs.
For example, both n-butane and isobutane have the same number of \chemmol{C} and
\chemmol{H} atoms (\chemmol{C\subscript{4}H\subscript{10}}), but have different chemical
properties due to different structure (see \ref{fig:organicegs}).
We will define this dissimilarity measure via a matching scheme
which attemps to match the atoms in one molecule to another.
The matching will only permit matching identical atoms,
i.e. carbon atoms can only
be matched to carbon atoms,
but we will incur penalties for matching atoms with
different bond types.

\insertFigOrganicEgs

\textbf{Molecules as graphs:}
For what follows, it will be convenient to view a molecule $M$ as a graph
$\molecule = (\atoms, \bonds)$, which is defined by a set of atoms $\atoms$ (vertices) and
a set of bonds $\bonds$ (edges). A bond $(u, v)\in\bonds$ is an unordered pair of atoms
$u, v\in\atoms$. Each atom $a\in\atoms$ has a label, denoted $\atomlabel(a)$, as does each
bond $b\in\bonds$, denoted $\bondlabel(b)$.
For example, $\atomlabel(a)$
could take values such as \chemmol{C}, \chemmol{H}, or \chemmol{O}, indicating carbon,
hydrogen, or oxygen atoms, while $\bondlabel(b)$ could take values such as 
\singlebond, \doublebond, or \aromaticbond{}, indicating single, double, or
aromatic bonds.
\ignore{In Figure~\ref{fig:organicegs}, atoms are represented by letters indicating their label; bonds are represented by lines where single lines indicate single bonds and
double lines indicate double bonds.}
We will also assign weights $\atomwt(a)>0$ for all atoms
$a\in\atoms$ of a molecule --
our matching scheme will attempt to match the weights in one molecule to another, in order
to compute a dissimilarity measure.
We will discuss choices for $\atomwt$ shortly.


\textbf{Description of the measure:}
Given two molecules $\molecule_1 = (\atoms_1, \bonds_1),
\molecule_2 = (\atoms_2, \bonds_2)$ with $n_1, n_2$ atoms respectively, 
let $U\in\RR_+^{n_1\times n_2}$ denote the matching matrix, i.e.
$U(i,j)$ is the weight matched between $i\in\molecule_1$ and
$j\in\molecule_2$.
The dissimilarity measure is the solution of the following program.
\begingroup
\allowdisplaybreaks
\begin{align*}
& \minimise_{U}\quad \atomtypepen(U) +
\structpen(U) + \nonmatchpen(U) 
\numberthis
\label{eqn:otdistdefn}
\\
\text{s.t.}\;
& \sum_{j\in\atoms_2} U(i,j) \leq \atomwt(i),\;
\sum_{i\in\atoms_1} U(i,j) \leq \atomwt(j),\; \forall i,j
\end{align*}
\endgroup
Here, the first term is the atom type penalty $\atomtypepen$ which only permits matching
similar atoms, i.e. $\chemmol{C}$ atoms can only be matched to other $\chemmol{C}$ atoms
and not $\chemmol{H}$ or $\chemmol{O}$ atoms.
Accordingly, it is defined as 

\[\atomtypepen(U) = \langle \atomtypeC, U \rangle
= \sum_{i\in\atoms_1} \sum_{j\in\atoms_2} \atomtypeC(i,j) U(i,j),\]

where $\atomtypeC(i,j) = 0$ if $\atomlabel(i) = \atomlabel(j)$ and $\infty$ otherwise.
The second term is the bond type penalty term, which, similar to $\atomtypepen$,
is given by $\structpen(U) = \langle \structC, U \rangle$,
where $\structC(i,j)$ is the penalty for matching unit weight from atom $i\in\atoms_1$ to atom $j\in\atoms_2$. We let $\structC(i,j)$ to be the fraction of dissimilar bonds in the union of all bonds. For example, in Figure~\ref{fig:organicegs}, $\structC(\atomone,\atomtwo) = 3/5$, since, between them they have one \chemmol{C-H} bond, three \chemmol{C-C}{} bonds, and one \chemmol{C=C} bond, of which the \chemmol{C-H} bond and one \chemmol{C-C} bond are common. If the atom type and bond type penalties are too large or infinite, we can choose to not match the atoms from
one molecule to another. However, we will incur a penalty 
via the non-matching penalty term $\nonmatchpen$. We set this term to be the sum of weights unassigned in both graphs, i.e. 
\begin{multline*}
\nonmatchpen(U) =
\sum_{i\in\atoms_1} (\atomwt(i) - \sum_{j\in\atoms_2} U(i,j)) \\ + \sum_{j\in\atoms_2} (\atomwt(j) - \sum_{i\in\atoms_1} U(i,j)).\end{multline*}
For two molecules $M_1, M_2$, we will denote the resulting dissimilarity measure,
i.e. the solution of~\eqref{eqn:otdistdefn}, by $\otdistsymbol$.

\textbf{Design choices:}
Let us first consider choices for the \textit{weights} 
$\{\atomwt(a)\}_{a\in\atoms}$ in the matching scheme.
A natural option
here is to let $\atomwt(a)$ be the atomic mass of atom $a$, which assigns more importance
to larger and heavier atoms, which heavily influence the 3D structure of the molecule.
Indeed, the molecular mass (sum of atomic masses)
is commonly used as an indicator of how drug-like a molecule
is in many metrics, including the QED~\citep{qed}.
However, lighter atoms may able to influence other important drug-like properties.
For example, the existence of hydroxyl groups (\chemmol{-OH}), is strongly correlated with
solubility in water, since it can function as an electron donor.
Hydrogen (atomic mass 1.008 Au)
plays a crucial role in this behaviour, and,
setting $\atomwt$ as above would downplay its significance
when compared to, say, carbon (atomic mass 12.011 Au).
In such cases, it is more appropriate to treat all atoms types equally,
setting  $\atomwt(a)=1$ for all atoms.

In addition to $\otdistsymbol$,
we also consider a \textit{normalized} version of this dissimilarity,
\[\otdistnormsymbol(M_1, M_2) = \otdistsymbol(M_1, M_2) / (\molwt(M_1) + \molwt(M_2))\]
where $\molwt(M) = \sum_{a\in A}\atomwt(a)$ is the total weight of a molecule $M=(A,B)$.
Our experience suggested that using $\otdistsymbol$ had a tendency to exaggerate
the dissimilarity between larger molecules, simply because a larger amount of atom weights
needed to be matched.
That said, the size of the molecule affects its drug-like properties (such as its ability
to bind with the target), and $\otdistsymbol$ accounts for the differences between small
and large molecules better than its normalized counterpart.

\textbf{Combining OT kernels:} The two options for the weights $\{\atomwt(a)\}_{a\in\atoms}$ and the two options
for normalization
give rise to four different combinations for our dissimilarity measure.
Instead of attempting to find a single best combination, we use an \textit{exponential sum} kernel
of the form $\kernel = e^{-\sum_i \beta_i d_i}$, where $\{d_i\}_i$ are the
measures obtained for each combination.
An ensemble approach of this form
allows us to account for all of the factors discussed above when comparing
molecules.
The $\beta_i$ terms, which affect the relative importance of each measure,
are treated as kernel hyperparameters which can be fitted using maximum likelihood
or posterior sampling.
It is worth mentioning that while the above form is similar to many popular
kernels, it is not known if it is in fact a valid positive definite kernel.
However, there are many ways to circumvent this issue in practice; in this work, we
project  the $n\times n$ matrix of $\kernel(\cdot,\cdot)$ values to the positive-definite
cone~\citep{kandasamy2018nasbot,sutherland2015scalable}.
In Appendix~\ref{sec:distdetails}, we show that~\eqref{eqn:otdistdefn} can be solved via
an optimal transport program~\citep{villani2008optimal} and discuss some shortcomings in
the proposed dissimilarity measure.

\textbf{A simple test:}
Finally, we perform a simple experiment to demonstrate that this dissimilarity metric
aligns with drug-like properties.
In Figure~\ref{fig:pairwise}, we provide the following scatter plot for molecules
sampled from the ChEMBL dataset.
Each point in the figure is for pair of networks. The x-axis
is the dissimilarity measure and the y-axis is the difference in the QED drug likeliness score~\citep{qed} and Synthetic accessibility score \citep{sascore}.
We use 100 molecules, giving rise to ~5000 pairs.
We see that when the measure is small, the difference in the QED score is close to 0.
As the measure increases, the points are more scattered.
One should
expect that for a meaningful distance measure, while molecules that are far apart could have either similar or different properties (as there could be several distinct ``clusters''),
molecules that are close by should have similar properties, and our measure satisfies this requirement.
Additionally, in Appendix~\ref{sec:distdetails}, we provide some interesting T-SNE visualizations
for our measure.

\insertPairwiseVisTwo

\subsection{Exploring the Space of Synthesizable Molecules and Optimizing the Acquisition}
\label{sec:explorer}

Our proposal for acquisition optimize involves randomly exploring the space of synthesizable
molecules and picking the one with the highest
acquisition---this can be viewed as
performing a random walk on a \emph{synthesis graph}\footnote{%
A synthesis graph is a directed graph where each node is a molecule, and the parents
of this node are the reagents, which when combined, produce the child molecule.
}.
For this, consider a setting in a laboratory or an automated experimentation apparatus,
where we have access to a limited library
of reagents $\Scal$ and process conditions $\Qcal$.
We will assume that we have access to an oracle \synthesize{} which can take as input
a set of compounds and process conditions and tell us the set of
molecules $M$ produced if these compounds are reacted in the given conditions.
In the event, a reaction cannot be effected, it will output \nullmol.
Our procedure for optimizing the acquisition function, described in 
Algorithm~\ref{algo:explorer}, operates as follows.
As input, it takes $\Scal$, $\Pcal$, the number of evaluations $n$ and a set
$D$ of evaluations where we have already conducted experiments.
First it randomly samples a few molecules $S$ and a few process conditions $Q$ from
$\Scal$ and $\Qcal$ respectively.
It passes them to \synthesize{} to generate a set of outputs $M$.
If the synthesis was successful, i.e. if we could generate new molecules that were
not evaluated before, they are added to the pool $\Scal$.
It repeats this for $n$ successful steps.
At the end, we return the maximizer $\argmax_{x\in\Scal}\varphi(x)$ of the
acquisition $\varphi$.

The above procedure relies crucially on the \synthesize{} oracle, which can
perfectly predict the outcomes of reactions.
Alas, no perfect such oracle exists\footnote{%
If it did,
the entire field of organic chemistry might be
expressed as a massive graph search problem.
}.
While outputs of reactions are well known for simple cases, it is impossible
to predict outcomes with complex molecules, and in some cases, the outputs may not even
be deterministic.
Fortunately however, there have been several advances in computational chemistry to
predict outcomes
of chemical reactions, which can be used in place of the oracle.
%
In our work we use Rexgen~\citep{coley-rexgen19}.
It should be emphasized
that since such predictors are not perfect, so in practice,
\Chem{} could end up recommending unsynthesizable molecules and/or incorrect
synthesis recipes.
An additional concern is that
the random walk in Algorithm~\ref{algo:explorer} could take
long and circuitous paths to arrive at a molecule. Consequently, the synthesis recipe
arrived at via Algorithm~\ref{algo:explorer}
may not be the most
efficient way to synthesize a given molecule.
Despite these concerns, we contend that our approach is far more likely to yield
synthesizable recommendations than existing approaches.
Developing synthesis predictors is an active area of
research~\citep{schwaller2018molecular,schwaller2018found}, and as such, as methods
become more reliable, so will the efficacy of our framework.
Moreover, an incorrect and/or inefficient recipe can still be a useful guide to a chemist
(who might choose to modify it), and in most cases
is better than expecting
the chemist to develop a recipe of her own from scratch.


\insertAlgoRandExplorer

%% file: experiments.tex
\section{Experiments}
\label{experiments}
\label{sec:experiments}

\textbf{Optimization objectives:}
We evaluate our methods on two of the most common molecular property functions found in
the literature: the QED score (Quantitative Estimate of Drug likeliness)~\citep{qed},
and Pen-logP score (penalized octanol-water partition coefficient).
The former is computed using the procedure described in~\citet{qed},
while the latter is computed using the following formula:
$\text{Pen-logP}(m) = \text{logP}(m) - \text{SA-Score}(m) - \text{ring-penalty}(m)$,
where logP is the octanol-water partition coefficient~\citep{miller1985relationships},
SA-Score is the synthetic accessibility score~\citep{sascore},
and ring penalty is the number of long cycles.
The partition coefficent measures solubility in water,
SA-score is a negative proxy for synthesizability (lower is easier), and
large rings might indicate that molecules are not stable once synthesized.
Note that the range of penalized logP is unbounded, and QED is constrained to values
between $0$ and  $1$.
In implementing Pen-logP, we followed the exact implementation of
this metric in \citep{jt-vae}.
We mention that these metrics may not be the most relevant to actual drug discovery applications -- for instance, they do not account for how well the molecule binds with 
the given target of interest.
However, their use in literature makes them good benchmarks to compare
different optimization methods.

\textbf{Methods:}
We compare three instantiations of \Chem{}: 1. using a molecular fingerprint kernel
(\fingerprint), 2. using the dissimilarity metric described in Section~\ref{sec:kernel}
(\dist), and 3. using a linear combination of \fingerprint{} and \dist{} (\sumkernel).
The fingerprint based kernel computes Tanimoto similarity between topological
(path-based) fingerprints of given molecules~\citep{rdkit}. The \sumkernel{} is a kernel given by $k(x, y) = \alpha_1\cdot\text{\fingerprint{}}(x,y) + \alpha_2\cdot \text{\dist}(x,y)$, where $\alpha_i\in [0,10]$ are kernel parameters fitted at training time. In addition, we also also compare to the random walk explorer (\rand) in
Algorithm~\ref{algo:explorer}, which operates exactly as described except returns
the maximum of the function $f$ in step~\ref{explorerreturn} (instead of the
acquisition).
This can be viewed as a simple random search baseline which attempts to optimize
in the space of synthesizable molecules.
We wish to reiterate that to our best knowledge other work do not enforce a
hard constraint on synthesizability, nor do they require that a recipe for synthesis
be provided. Hence, they are not directly comparable to our method.
However, we quote results on the best QED and Pen-logP values from their papers for
comparison. Moreover, we include an additional virtual screening baseline, which is allowed to randomly sample and evaluate molecules from the entire dataset, instead of just the compounds reachable by synthesis from the starting pool.

\textbf{Experimental set up:}
As stated previously, we wish to emulate a setting where a chemist has to work with the
reagents and process conditions available to her.
We choose 20 randomly chosen molecules from the openly available
ChEMBL database as our initial set of reagents.
The maximum QED score of the initial pool was $0.858$ (when
QED > 0.9, it is typically considered high).
As the process conditions for the random explorer,
we use all the process conditions available in Rexgen.
We bootstrap all three methods listed above by evaluating the metric (QED or Pen-logP)
on this initial set, and then execute the methods for $80$ iterations, totaling 100
evaluations of $f$.
We describe additional details on our BO implementation in
Appendix~\ref{sec:appexperiments}.


\insertOptimGraphs

\vspace{-2mm}
\subsection{Results \& Discussion}

\textit{Main Results:} In Figure~\ref{fig:optim_graphs}, we plot the number of iterations against the
optimal found value by each method over $80$ function evaluations for both
QED and Pen-logP.
We provide the final optimal values for each method in Table~\ref{tab:overall_res}.
The results were obtained by averaging over $5$ independent runs.
\Chem{} methods, \fingerprint{}, \dist{} and \sumkernel{}, all outperform the naive random walk
strategy on both tasks, validating the use of model based Bayesian strategies
for this task.
\dist{} does better than \fingerprint{} on the QED score while vice versa
on Pen-logP, and \sumkernel{} provides a good adaptive trade-off between them that works well for both benchmarks, and also has lower variance.
\ignore{It is worth mentioning that, generally speaking, 
the QED score is considered a more holistic view of drug likeliness
than Pen-logP~\citep{qed}.}

\emph{Optimal Molecules \& Synthesis Recipes:}
Figure~\ref{fig:molecule_pics} illustrate some optimal molecules found
for the QED and Pen-logP objectives by \Chem.
For the most part,
optimal QED molecules were found by \dist, while optimal Pen-LogP molecules
 by \fingerprint.
Interestingly, molecules with high QED scores tend to be simpler
than those with high Pen-logP scores.
In Appendix~\ref{sec:appexperiments}, we visualize and discuss the synthesis recipes
for some of the optimal molecules.

\insertImprovementTable

\emph{Reliability of synthesis paths:} A thorough validation of the synthesis paths proposed by \Chem{} would require performing actual synthesis in lab conditions. However, we can perform the following sanity checks. Using synthetic accessibility score \cite{sascore} as a proxy for ease of synthesis of the resulting molecule, we can evaluate plausibility of the end result. The results presented in Table~\ref{table:sas_scores} show that the end molecule is within a reasonable range from averages in curated datasets ChEMBL and ZINC, and the minimum score over synthesis path is well below these values.

\insertSASTable

\emph{Novel Molecules:} 
During the execution of \Chem{}, we compute the
fraction of molecules that do not appear in the entire ChEMBL dataset.
For \dist{} optimizing QED, on average $95.64\%$ molecules are novel,  for \fingerprint{} $96.84\%$; and for Pen-logP $78\%$ and $87.67\%$, respectively.
This indicates that \Chem{} is able to explore the chemical space well, despite the
constraints on synthesizability.

\textbf{Comparison with existing work:} In Table~\ref{tab:compar_res}, we compare
 \Chem{} to state-of-the-art
methods adopting reinforcement learning or generative modeling
techniques~\citep{organ,jt-vae,gcpn,zhou2018optimization}.
We use the same evaluation
strategy as in these works, reporting top scores across several runs.
It is interesting to compare the number of QED/Pen-LogP evaluations required by some
of these methods.
\citet{organ} is trained with supervision on a random
subset of $5K$ molecules from the ZINC dataset~\citep{zinc},
and hence uses at least $5K$ evaluations. VAE in
\cite{jt-vae} is trained on full ZINC dataset ($\approx 250k$ molecules) in an
unsupervised manner, and then $25k$ evaluations to train a GP and optimize the given objective.
Both \cite{gcpn} and \cite{zhou2018optimization}
train RL policies using all the $250K$ molecules in the ZINC dataset and incorporate
the penalized logP or QED
score as part of the reward, hence making at least that many evaluations. 
In contrast, in our \Chem{} experiments, we ran $100$ BO iterations using two different
kernels for $5$ trials, totalling $1000$ function evaluations.
It should be emphasised that the above methods are not designed to keep the number of
QED/Pen-logP evaluations to a minimum, and in fact, are tools developed for very
different settings.
Yet, it speaks to the efficiency of \Chem{}, that we were able to obtain better or
comparable values than the above work in significantly fewer evaluations,
particularly given our more stringent conditions on synthesizability.

\insertComparisonTable
\insertMoleculePics

\textbf{Virtual screening baseline:} There are two possible ways to translate the virtual screening experiment into a computational simulation. In the first version, we assume that a fixed number of compounds is available to an experimenter (same as the starting pool), and we can either synthesize from them, or use them directly for screening. This baseline is already part of the results above, since we spend the initial BO budget on finding the maximum value of the initial pool (``screening'' it), and after that all optimizers have to improve upon that value. In the second version, we compare virtual screening and \Chem{}/\rand{} for the same number of evaluations. Now we start with a pool, and then sample compounds \textit{outside} of that pool from the rest of the dataset. This corresponds to a situation where the experimenter purchases the compounds randomly in addition to the ones she has; in theory, this could lead to a larger optimum due to accessing more dataset molecules than in our setup, i.e. a larger search space. The results obtained by simulating such an experiment are shown in Figure \ref{table:virt}. Even with using more samples, these values are worse than the numbers in Table \ref{tab:overall_res}.

\insertVirtualScreeningTable



%% file: conclusion.tex
\section{Conclusion}\label{disc}
\label{sec:conclusion}

In real world use cases for computational and statistical methods for molecular
optimization,
an algorithm recommends a molecule,
which is synthesized, tested, and the results returned to the algorithm.
These results are then used by the algorithm to inform future recommendations.
In order to achieve full automation,
computational methods should strive to ensure that such recommendations
are synthesizable and provide a recipe to do so.
\Chem{}, which uses BO techniques to design recommendations, is a first step towards
this ambitious goal.
Our experiments indicate that model-based Bayesian methods can outperform naive
alternatives for this problem. We study kernels for \Chem{} and find that the \dist{} kernel we propose can outperform standard kernels in some tasks, and that combining it with other kernels (such as \fingerprint{}) can be a lower-variance alternative that performs well across benchmarks.
In addition, on two benchmark objectives, we are able to get competitive or better scores than existing work, while using significantly less evaluations of the objective.
While our approach is invariably constrained by limitations of
current synthesis predictors,
it can still be a very useful guide to a practitioner.

Improving the reliability of synthesis predictors and  developing smarter methods to
explore the chemical space are interesting avenues for future research, which will
improve the efficacy of our framework.
Another direction is to use \Chem{} (and other methods)
to optimize for the ability to bind with a given target.
Separately,
it would also be interesting to view the optimization  budget 
not in terms of the number of compounds tested, but rather in terms of  the number of additional
synthesis steps required (it is plausible that synthesis is the bottleneck, not the cost of testing).
This paradigm also brings up some new interesting methodological questions for Bayesian optimization.
Finally, it would be interesting to extend and test our framework on biologics and
other molecular optimization problems in drug discovery and materials science.

%% file: app_dist.tex
\section{Some Additional Details on the Dissimilarity Measure}
\label{sec:distdetails}

\subsection{Solving~\eqref{eqn:otdistdefn}}
\label{sec:distsolving}

In this section, we describe how the linear program for computing the dissimilarity
measure~\eqref{eqn:otdistdefn}, can be solved using an optimal transport (OT)
program~\citep{villani2008optimal}.
This reformulation is similar to that of~\citet{kandasamy2018nasbot}, who use OT
to describe a distance between neural network architectures.

Say we are given two molecules $\molecule_1 = (\atoms_1, \bonds_1),
\molecule_2 = (\atoms_2, \bonds_2)$ with $n_1, n_2$ atoms respectively,
let $U\in\RR_+^{n_1\times n_2}$ denote the matching matrix, i.e.
$U(i,j)$ is the weight matched between $i\in\molecule_1$ and
$j\in\molecule_2$.
We now define a sequence of variables which form the parameters of our OT program.
First, let $\molwt(M_i) = \sum_{a\in A_i}\atomwt(a)$ is the total weight of a molecule
$M_i=(A_i,B_i)$ for $i=1,2$.
Denote $y_1 = [\{\atomwt(a)\}_{a\in A_1}, \molwt(M_2)] \in \RR^{n_1 + 1}$ and
$y_2 = [\{\atomwt(a)\}_{a\in A_2}, \molwt(M_1)] \in \RR^{n_2 + 1}$.
Next, let $C = \atomtypeC + \structC \in \RR^{n_1\times n_2}$
and $C' = [C \, \one_{n_1}; \one_{n_2}^\top \, 0] \in  \RR^{(n_1+1)\times (n_2+1)}$;
i.e. $C'$ has $\atomtypeC+\structC$ in its first $n_1\times n_2$ block, representing
the atom type and bond type penalties in~\eqref{eqn:otdistdefn}, while
the $1$'s in the last row and column capture the non-matching penalty.
We finally let $U'\in \RR^{(n_1+1)\times (n_2+1)}$ be our optimization variable where
the first  $n_1\times n_2$ block will correspond to the optimization variable
$U$ in the original program.
It is easy to see that~\eqref{eqn:otdistdefn} is equivalent to the following linear program,
which is an optimal transport program:
\begingroup
\allowdisplaybreaks
\begin{align*}
& \minimise\quad \langle U', C' \rangle
\\
& \subto\quad
U'\one_{n_2+1} = y_1,
\hspace{0.1in}
U'^\top\one_{n_1+1} = y_2.
\end{align*}
\endgroup
We refer the reader to 
Theorem 2 in~\citet{kandasamy2018nasbot}, who formally prove this result in a similar
setting.

\subsection{T-SNE visualizations for the OT distance}
\label{subsec:tsne_vis}

We perform another experiment to verify the validity of the proposed optimal transport
dissimilarity measure. We use the four
different base combinations of settings for the OT distance to compute distances between
$200$ randomly sampled molecules, and use these distances to compute 2-dimensional t-SNE embeddings \citep{tsne}.
These embeddings aim to preserve distances, so that visual
closeness translates into OT-distance closeness.
We also color the points by values of QED
(drug-likeliness) and synthetic accessibiility scores. The results are shown in Figure
\ref{fig:tsne_vis}. We see that despite the fact that the chemical space has complicated
dependencies between molecule structure and properties, dependencies in the induced
embedding space are relatively continuous.
We can also observe clusters of molecules with similar values. In Figure \ref{fig:tsne_vis2}, we compare the planar embeddings produced by other possible distances: $\ell_2$ distance between pairs of fingerprints and inverted Tanimoto similarity measure  between molecules (referred to as \fingerprint{} kernel in the main part of the paper), one may say OT-dist looks slightly better (e.g. low versus high 
values are more separated in the plots).

\subsection{Some Known Limitations}
\label{sec:distshortcomings}

\textbf{Stereoisomers:}
Since our dissimilarity measure is based on the graph representation, it will not be able
to distinguish between stereoisomers, i.e. molecules which have the same formula and
bonded atoms, but different 3D orientation.
For example, pictured below are D-Glucose and L-Glucose.
Since, they have the same graph representation, our dissimilarity measure will be $0$
between both molecules.
However, they have different 3D structures (being mirror images of each other),
which can give rise to different physical properties.
For instance, D-Glucose can be digested by the human body while L-Glucose cannot.

\insertFigGlucose

It is worth noting that many graph convolution based approaches for modeling molecules
face this challenge.
One way to circumvent this issue is to combine our kernel with other features which
account for 3D structure in a sum or product kernel.

%% file: app_model_details.tex
\section{Some Implementation Details}
\label{sec:model_details}
For the BO methods, we fit GP hyperparameters by maximizing the marginal likelihood.
As the acquisition, we adopt the ensemble method described in~\citep{dragonfly} using
the EI, UCB, and TTEI acquisitions instead of sticking to a single acquisition.
To optimize the acquisition, we ran the explorer for $20$ iterations on each BO iteration,
but added the new molecules to our initial pool $\Scal$ for the next iterations,
so that we can search across a large pool during the entire optimization routine. This corresponds to ``reusing'' explored and synthesized compounds in a real experiment.

%% file: app_experiments.tex
\section{Additional Experimental Results}
\label{sec:appexperiments}

\textbf{Experiments with low starting value}

To verify that \Chem{} successfully optimizes the objective regardless of the quality of initial pool, we conduct an experiment on pools of 20 molecules randomly selected from subset of ChEMBL dataset that has value of the objective function capped by $0.7$ for QED and $3$ for penalized LogP function (approximately 60\% percentiles in ChEMBL). The results below show that \Chem{} performs well in such cases, too, and does so better than baseline with the same regularities as before (the \fingerprint{} kernel performs worse than \dist{} kernel on QED and better on penalized LogP task).

\insertOptimGraphsLowStart

\textbf{Synthesis Paths}

We visualize the synthesis paths for some of the optimal molecules in
Figures~\ref{fig:syn_path1}-\ref{fig:syn_path4}.
The boxed molecules are from the initial pool of $20$ reagents.
In this figure, when arrows from two or more parent molecules point to a child
molecule, it means that the child molecule was obtained by reacting the
parent molecules.

It is worth mentioning some caveats here.
First,
we see a few cases of complex molecules being combined to produce a simpler
molecule -- the most striking example being the one in Figure~\ref{fig:syn_path2}
where two complex molecules are combined to produce Methane
(\chemmol{CH\sbsc{4}})\footnote{%
In reality, Methane was probably just meant to be a by-product of a reaction meant
to produce some other molecule.
}.
It is more likely that simpler molecules will be available as reagents in a realistic
setting.
This is an artefact of our initial pool, and we believe that such cases can be
avoided by carefully selecting an initial pool.
Second, note that in all synthesis paths shown, there are molecules with large rings.
Large  rings are not necessarily stable, and hence such molecules are hard to synthesize.
We believe this could be due Rexgen, and, as mentioned in the main text, when
such synthesis predictors become more accurate and reliable, so will the
efficacy of our proposed framework.

The red boxes in the molecules are because RDkit's 2D layout algorithm overlays two
atoms -- which is likely to happen with large molecules.

\textbf{Some statistics on the ChEMBL Dataset:}
In Figure~\ref{fig:data_stats}, we plot the distribution of QED and Pen-logP on
the ChEMBL dataset. These values help us understand the success of optimization procedures relative to the average over the dataset from which the starting pool was drawn: the histograms show that the optimized values lie in the highest percentiles of the original dataset.



\insertDataStats

\insertSynPaths

\insertTsneVis

\insertTsneVisDifferent